\documentclass{opendatalab}

\usepackage[T1]{fontenc}
\usepackage{hyperref}
\usepackage{url}
\usepackage{booktabs}
\usepackage{amssymb}
\usepackage{graphicx}
\usepackage{wrapfig}
\usepackage{enumitem}
\usepackage{array}
\usepackage{makecell}
\usepackage{multirow}
\usepackage{xcolor}
\usepackage{siunitx}
\usepackage{colortbl}
\usepackage{tabularx}
\usepackage{listings}
\usepackage[most]{tcolorbox}
\usepackage{titletoc}

\providecommand{\benchyes}{\ensuremath{\checkmark}}
\providecommand{\benchno}{\ensuremath{\times}}
\providecommand{\benchpartial}{\ensuremath{\triangle}}
\newcommand{\benchmarkhead}[2]{%
    \makecell[c]{%
        \mbox{\textbf{#1}}\\[-0.15em]
        {\tiny #2}%
    }%
}

\definecolor{codegreen}{rgb}{0,0.6,0}
\definecolor{codegray}{rgb}{0.5,0.5,0.5}
\definecolor{codepurple}{rgb}{0.58,0,0.82}
\definecolor{backcolour}{rgb}{0.95,0.95,0.92}
\definecolor{promptcolor}{HTML}{D1D0F2}
\definecolor{promptcolorheader}{HTML}{bdbcec}

\definecolor{promptcolor}{HTML}{E3F0FA}
\definecolor{promptcolorheader}{HTML}{B5D6ED}
\definecolor{prompttitletext}{HTML}{1B3A5C}

\newtcolorbox{promptbox}[1][]{
	enhanced, breakable,
	top=0.3em,bottom=0.3em,left=0.5em,right=0.5em,
	toptitle=0.3em,bottomtitle=0.2em,boxsep=0pt,
	colframe=promptcolorheader, colback=promptcolor!50, boxrule=0.5pt,
	width=\columnwidth, 
	coltitle=prompttitletext,
	title={\footnotesize #1} 
}
\lstdefinestyle{promptstyle}{
    backgroundcolor=\color{backcolour},   
    commentstyle=\color{codegreen},
    keywordstyle=\color{magenta},
    numberstyle=\tiny\color{codegray},
    stringstyle=\color{codepurple},
    basicstyle=\ttfamily\footnotesize,
    breakatwhitespace=false,         
    breaklines=true,                 
    captionpos=b,                    
    keepspaces=true,                 
    numbers=left,                    
    numbersep=5pt,                  
    showspaces=false,                
    showstringspaces=false,
    showtabs=false,                  
    tabsize=2
}
\title{Beyond Endpoint Performance: Process-Level Evaluation of Self-Evolving Agents}
\author[1]{Hongqiang Lin$^*$}
\author[2]{Chao Liu$^*$}
\author[2]{Xiaofan Bai}
\author[2]{Xuan Jin}
\author[2]{Yuhong Li}
\author[1]{Nenggan Zheng}
\author[2]{Xipeng Cao}

\affiliation[1]{Zhejiang University}
\affiliation[2]{Alibaba Group}

\abstract{
Self-evolving agents convert interaction feedback into persistent artifacts, such as memories or skills, which in turn guide subsequent decisions. As these artifacts are iteratively updated throughout an experience stream, the capabilities they support may evolve. Consequently, endpoint performance alone offers an incomplete view of self-evolution. Process-level evaluation is therefore essential to identify when a target capability emerges and whether later updates strengthen, preserve, or weaken it. Motivated by this, we propose \textsc{EvoPathBench}, a benchmark that tracks individual capabilities during artifact-level self-evolution. EvoPathBench fixes the base model, tools, freezes evolving artifacts at successive checkpoints, and evaluates the target capability on held-out episodes. This benchmark evaluates agent self-evolution using public trading data and calibrated trajectories. It tests three capabilities: generalization to unseen tasks, retention after unrelated learning, and rule adaptation to new evidence. Experimental results show that gains on similar unseen tasks often weaken under distribution shift, retention losses are concentrated in a minority of evolution paths, and no method achieves reliable rule adaptation. Moreover, while self-evolution enables agents to generate candidate artifacts with substantial held-out gains, the selected updates consistently fall short of realizing this potential. Together, these findings establish capability-level process evaluation as a foundation for analyzing self-evolution, identifying candidate evaluation and selection as key targets for improvement.
}

\date{\today}
\correspondence{Nenggan Zheng (\email{zng@cs.zju.edu.cn}), Xipeng Cao (\email{caoxipeng.cxp@alibaba-inc.com}).
}

\metadata[Code]{\url{https://github.com/HQ-Lin/EvoPathBench}}
\begin{document}

\maketitle

\section{Introduction}
Self-evolving agents engage in continuous environmental interaction,
translating feedback from action outcomes into updates of persistent memories,
skills, or workflows \cite{zhang2025aflow,xue2026rethinkingselfevolvingagentsneed,ouyang2026skilloslearningskillcuration,lin2026rethinkingselfevolutionconstrainedexplorationexploitation}. By updating these persistent artifacts, agents carry knowledge from earlier interactions into later tasks and improve their behavior as experience accumulates. This ability offers a practical path toward agents that continue to learn during extended deployment \cite{lin2026museautoskillselfevolvingagentsskill}. However, these updates are useful only if the resulting capabilities remain effective as the agent encounters new tasks and continues to learn. Evaluating how these capabilities form and evolve over time is therefore essential for determining whether self-evolution produces meaningful and reliable improvement \cite{xu2026evoarenatrackingmemoryevolution, xue2026pastbenchbenchmarkingfoundationsrecursive}.

Self-evolution is inherently sequential and path dependent. Each update modifies the persistent artifact state available to the agent, which may in turn influence both its subsequent decisions and its response to new experience \cite{mao2026practicemakesunsafeskill,tang2026wikiskillcompilingagentexperience,bai2026skillzipproexecutionawaredynamic,jiang2026seaevalbenchmarkevaluatingselfevolving}. Evaluating this process requires two key elements. First, learning episodes must be organized as an ordered task stream rather than treated as samples from a static dataset. This stream dictates what the agent learns, when it learns it, and the sequence in which its artifact state is updated. Without this ordering, changes in later behavior cannot be attributed to prior experience \cite{yan2026agentstreamselfevolvingllmagents,gao2026tmlr-survey}. Second, preserving artifact states across the stream is insufficient if evaluation relies solely on an aggregate endpoint score, as improvements on newly learned tasks may mask degradation in previously acquired capabilities. Therefore, the same focal capability must be assessed at intermediate checkpoints both before and after subsequent artifact updates. This capability-level process evaluation reveals when the capability first appears and whether subsequent updates strengthen, preserve, or weaken it.

Standard static benchmarks reset the agent’s state before each task and evaluate tasks independently \cite{jimenez2024swebenchlanguagemodelsresolve,patil2025the,he2026livemathematicianbench,xie2024osworldbenchmarkingmultimodalagents}. As a result, they only capture a snapshot of current performance rather than tracking evolution progress over time. Recent benchmarks satisfy the first requirement by preserving agent artifact state across ordered task sequences, yet they fail to meet the second \cite{zhong2026skilllearnbenchbenchmarkingcontinuallearning,xue2026pastbenchbenchmarkingfoundationsrecursive,xu2026evoarenatrackingmemoryevolution,zheng2026seagymevaluationenvironmentselfevolving}. These benchmarks typically aggregate performance across multiple tasks or capabilities at each checkpoint. Such aggregate scores can obscure changes in individual capabilities. As a result, these evaluations indicate whether the agent improves globally but fail to reveal how a previously learned capability evolves under subsequent artifact updates.

This gap motivates capability-level process evaluation along three complementary dimensions. \emph{Learning generalization} measures whether a capability acquired from observed episodes applies to unseen episodes. \emph{Capability retention} measures whether that capability remains available as the agent continues to learn. \emph{Rule adaptation} measures whether the agent revises an acquired rule when new evidence shows that the underlying relation has changed. Together, these dimensions trace how a
capability forms, persists, and changes during self-evolution.

We present \textsc{EvoPathBench} to support this evaluation for artifact-level self-evolution. The benchmark holds the base model, tools, and executor fixed while persistent memories, skills, or workflows evolve. At fixed checkpoints, it freezes these artifacts and reevaluates the same target capability on held-out episodes excluded from subsequent updates. Three separate task-stream templates expose the agent to repeated evidence, unrelated learning, or a changed relation, allowing the corresponding capability changes to be measured independently. Our contributions are as follows:

\begin{itemize}
    \item \textbf{Capability-level process evaluation.} We formalize self-evolution as persistent artifact updates and introduce metrics to evaluate agent capability dynamics during this process.
    \item \textbf{Benchmark design.} EvoPathBench contains 1,800 streams across six task families and two market layers, combining public trading data with calibrated generated trajectories. These streams are organized into \emph{Accumulation}, \emph{Interference}, and \emph{Reversal} categories to evaluate learning generalization, capability retention, and rule adaptation, respectively.

    \item \textbf{Empirical findings.} Experimental results show that gains diminish under distribution shift, forgetting is concentrated in a minority of paths, and reliable rule adaptation remains unestablished, indicating that no method performs consistently across all three dimensions.
    
   \item \textbf{Analysis of skill evolution.} We examine artifact growth, token usage, cross-model transfer, candidate search, and acceptance safeguards. Evolved skills achieve SFT-like gains, yet additional computation does not guarantee better evolution, with candidate evaluation identified as the primary bottleneck.
\end{itemize}

\section{Related Work}
\subsection{Self-Evolving Agent}
The development of self-evolving LLM agents marks a progression from temporary adaptation to persistent self-modification \cite{zweiger2025selfadapting,zhang2026darwin}. Early reasoning and reflection methods used feedback to refine agent behavior within a single task or over repeated attempts, but generally did not persist these adaptations for reuse in subsequent tasks \cite{yao2023react,shinn2023reflexion}. Subsequent experience-learning methods introduced cross-task persistence through interaction-derived memories, evolving contexts, and executable skill libraries \cite{wang2024voyager,chhikara2025mem0buildingproductionreadyai,zhang2026agentic}. These persistent components, which we call evolution artifact states, allow past experience to influence future behavior while the base model remains fixed. More recent systems automate the artifact lifecycle through skill discovery, validation, curation, and revision\cite{xia2026skillrlevolvingagentsrecursive,zhang2026coevoskills,lin2026rethinkingselfevolutionconstrainedexplorationexploitation,ni2026trace2skilldistilltrajectorylocallessons,yang2026skilloptexecutivestrategyselfevolving,wang2026skillx}. This progression gives rise to artifact-level self-evolution, in which experience from one episode modifies persistent agent state used in later episodes. However, such evolution is not necessarily monotonic: newly acquired artifacts may interfere with previously established capabilities, and unsafe rules may persist beyond their originating episodes \cite{mao2026practicemakesunsafeskill}. EvoPathBench evaluates how these artifacts are acquired, transferred, retained, and forgotten over ordered task streams, while keeping the base model, tools, and executor fixed.

\subsection{Benchmarks for Longitudinal Agent Evaluation}
\label{sec:related-longitudinal-evaluation}
Standard agent benchmarks treat tasks as a static dataset rather than an ordered learning stream. Because agent state is reset between episodes, they measure current performance rather than evolution over time
\cite{jimenez2024swebenchlanguagemodelsresolve,patil2025the,xie2024osworldbenchmarkingmultimodalagents,he2026livemathematicianbench}.
Recent benchmarks address this limitation by preserving artifact state across ordered task streams \cite{wei2026evomemorybenchmarkingllmagent,zhong2026skilllearnbenchbenchmarkingcontinuallearning,xue2026pastbenchbenchmarkingfoundationsrecursive,xu2026evoarenatrackingmemoryevolution,zheng2026seagymevaluationenvironmentselfevolving,deng2026finevobenchlongitudinalbenchmarkselfevolving}.  As Table~\ref{tab:benchmark-dimensions} indicates, existing benchmarks do not fully support capability-level process evaluation. Such evaluation requires measuring the same focal capability on held-out episodes both before and after specific artifact updates.  This process-level evaluation reveals when the capability emerges and whether later updates preserve, weaken, or revise it, whereas scores aggregated across tasks may hide these changes. Moreover, these benchmarks rely on existing datasets, curated environments, or real-case workflows, which rarely combine realistic market behavior with controlled task construction. In contrast, EvoPathBench operationalizes capability-level process evaluation through three distinct task-stream templates: \emph{Accumulation} for learning generalization, \emph{Interference} for capability retention, and \emph{Reversal} for rule adaptation. Its hybrid design combines calibrated real-market trading data with generated data that preserve similar market properties, providing both realistic market dynamics and controlled changes to task relations.

\begin{table}[t]
\centering
\caption{Comparison of representative longitudinal agent benchmarks. \benchyes{} indicates that the dimension is explicitly evaluated, \benchpartial{} indicates that only part of the dimension is evaluated, and \benchno{} indicates that the dimension is not evaluated. \emph{Existing} denotes previously released datasets or benchmark tasks. \emph{Curated} denotes tasks or environments assembled under predefined rules. \emph{Real} denotes workflows derived from real cases. \emph{Hybrid} denotes calibrated real-market trading data combined with generated data that preserve similar market properties.}
\label{tab:benchmark-dimensions}
\begingroup
\setlength{\tabcolsep}{3.0pt}
\renewcommand{\arraystretch}{1.15}
\fontsize{7.5pt}{8.8pt}\selectfont
\resizebox{\linewidth}{!}{%
\begin{tabular}{
    @{}
    l
    *{6}{c}
    >{\columncolor[gray]{0.93}[\tabcolsep][0pt]}c
    @{}
}
\toprule

\textbf{Evaluation design}
& \benchmarkhead{SkillLearnBench}{
    \citealp{zhong2026skilllearnbenchbenchmarkingcontinuallearning}
}
& \benchmarkhead{Evo-Memory}{
    \citealp{wei2026evomemorybenchmarkingllmagent}
}
& \benchmarkhead{PAST-Bench}{
    \citealp{xue2026pastbenchbenchmarkingfoundationsrecursive}
}
& \benchmarkhead{Evo-Arena}{
    \citealp{xu2026evoarenatrackingmemoryevolution}
}
& \benchmarkhead{SEA-Gym}{
    \citealp{zheng2026seagymevaluationenvironmentselfevolving}
}
& \benchmarkhead{FinEvo-Bench}{
    \citealp{deng2026finevobenchlongitudinalbenchmarkselfevolving}
}
& \benchmarkhead{EvoPathBench}{
    \textit{Ours}
}
\\

\midrule

Ordered task stream
& \benchno
& \benchyes
& \benchyes
& \benchyes
& \benchyes
& \benchyes
& \benchyes
\\

Learning generalization
& \benchyes
& \benchpartial
& \benchyes
& \benchpartial
& \benchyes
& \benchpartial
& \benchyes
\\

Capability retention
& \benchno
& \benchno
& \benchpartial
& \benchpartial
& \benchyes
& \benchno
& \benchyes
\\

Rule adaptation
& \benchno
& \benchno
& \benchyes
& \benchyes
& \benchpartial
& \benchno
& \benchyes
\\

\midrule

Data source
& Existing
& Existing
& Curated
& Curated
& Existing
& Real
& \textbf{Hybrid}
\\
\bottomrule
\end{tabular}%
}
\endgroup
\vspace{-0.6em}
\end{table}

\section{EvoPathBench}
\label{sec:evopathbench}

EvoPathBench evaluates three capabilities of artifact-level self-evolution: \emph{learning generalization}, \emph{capability retention}, and \emph{rule adaptation}. Learning generalization asks whether an agent can learn from a small number of episodes and apply what it learns to new episodes from the same task family, rather than fit only the observed cases. Capability retention asks whether the agent can continue learning without losing capabilities acquired earlier. Rule adaptation asks whether the agent can recognize that an old rule no longer applies and revise it using new evidence. EvoPathBench evaluates these capabilities through accumulation, interference, and reversal streams, using matched hidden probes to test frozen artifact states at successive checkpoints.

\subsection{Benchmark Overview}
\label{sec:benchmark-overview}

\paragraph{Artifact-level evolution.}
We hold the base model, tools, and executor fixed for each method. Evolution is restricted to explicit artifact states, such as memories, skills, and workflow rules. Let $a \in \mathcal{A}$ denote a self-evolution method, let $\pi_\theta$ denote an LLM with fixed parameters $\theta$, and let $m_{c,t}^{a}$ denote the artifact state after episode $t$ along evolution path $c$. Executing episode $e_{c,t}$ from $m_{c,t-1}^{a}$ produces a trace $z_{c,t}^{a}$ containing the actions, observations, and feedback available to the agent. The state evolves according to
\begin{equation}
m_{c,0}^{a}=m_0^{a},\qquad m_{c,t}^{a}=U_a\!\left(m_{c,t-1}^{a},z_{c,t}^{a},\omega_{c,t}^{a}\right),
\label{eq:self-evolution}
\end{equation}
where $U_a$ is the artifact update rule and $\omega_{c,t}^{a}$ captures update randomness. Future tasks and held-out evaluation outcomes are unavailable to $U_a$. The object of evaluation is therefore the evolving process $\mathcal E_a=(m_{c,0:t}^a,\pi_\theta,U_a)$ rather than a single final policy.

\paragraph{Task families and streams.}
Let $\mathcal F$ denote the set of task families. For each family $f\in\mathcal F$, $\mathcal L_f$ contains update-enabled learning episodes and $\mathcal H_f$ contains held-out evaluation episodes.  Episodes in the two sets share the same underlying decision problem but differ in market windows, parameters, observation mappings, or opponents. Episodes in $\mathcal H_f$ are used only for evaluation and cannot update the agent's artifacts. An evolution path is an ordered stream
\begin{equation}
\mathcal S_c=\bigl((f_{c,1},e_{c,1}),\ldots,(f_{c,T},e_{c,T})\bigr),\qquad e_{c,t}\in\mathcal L_{f_{c,t}}.
\label{eq:task-stream}
\end{equation}
Tasks arrive one at a time and future tasks are hidden. Episode-local interaction history is cleared after each task, while the permitted persistent artifacts remain available.

EvoPathBench combines public real trading data from Binance Spot with synthetic data generated from the market patterns observed in these records. This design preserves realistic market dynamics while supporting controlled task construction. EvoPathBench contains six controlled families. \emph{Trend} introduces persistent drift. \emph{Mean reversion} creates temporary price deviations that tend to decay. \emph{Liquidity} increases execution friction. \emph{Event jump} introduces abrupt heavy-tailed shocks. \emph{Opponent reflexivity} strengthens the effect of recent order flow and other agents on subsequent prices. \emph{Risk contract} combines volatile market conditions with binding limits on position, order size, turnover, drawdown, and short selling. Further details are provided in Appendix~\ref{app:benchmark-composition}.

Each stream contains five update opportunities and follows one of three templates, with each template corresponding to one evaluation dimension. The \emph{Accumulation} template presents five learning episodes from one focal family, where the focal family is the family whose capability is being tracked. The \emph{Interference} template presents two episodes from the focal family, followed by three episodes from distinct distractor families. The \emph{Reversal} template presents three episodes under an initial relation, followed by two counterevidence episodes in which that relation changes.

Streams are instantiated in two environment layers. In the \emph{exogenous} layer, the market path is independent of the focal agent's actions. In the \emph{endogenous} layer, the focal agent trades with five fixed opponents and order imbalance affects later prices. The release contains 1,800 balanced longitudinal streams across three templates, six task families, and two market layers.

Different held-out evaluation episodes support these measurements. Near-distribution and transfer episodes evaluate learning generalization. Retention episodes are repeated across checkpoints to measure capability loss. Post-reversal episodes evaluate behavior under the changed relation. 

\subsection{Evaluation Dimensions}
\label{sec:evaluation-dimensions}
Each stream yields six artifact states, denoted by $K_0,\ldots,K_5$. Let $k\in\{0,\ldots,5\}$ index these states: $k=0$ precedes all updates, and each
$k\geq1$ follows the $k$-th update opportunity. At every $k$, we freeze the artifact state and evaluate a read-only copy on matched held-out episodes from
$\mathcal H_f$. Crucially, this evaluation is completely isolated from the training process. Let $Y^{a}_{c,k,e,r}$ denote the evaluation score for method $a$ in evolution path $c$, evaluated at state index $k$ on held-out episode $e$ in execution repeat $r$. While repeats preserve the artifact state and environment realization, they trigger an independent model execution to isolate inference-time stochasticity.

\paragraph{Attributing performance changes.}
An increase across checkpoints alone does not prove self-evolution. Therefore, we use two paired comparisons to isolate the evolution procedure's effect from the artifact's effect. First, we compare the evolving method against a non-evolving \emph{Baseline} (which follows the same stream without updating artifacts) to measure the total evolution effect. Second, to test the artifact's direct influence, we apply a \emph{state-off} intervention (disabling artifact access during testing). This is an evaluation-time manipulation, not a separate evolution path.

Let $Y^{a,\mathrm{off}}_{c,k,e,r}$ denote the state-off score. Using the same evolution path $c$, checkpoint $k$, held-out episode $e$, and execution repeat $r$ in each pair, we define the \emph{capability evolution gain} (CEG) and the \emph{artifact state-use effect} (SUE) as
\begin{align}
\mathrm{CEG}^{a}_{k,f}&=\mathbb{E}\!\left[Y^{a}_{c,k,e,r}-Y^{\mathrm{baseline}}_{c,k,e,r}\mid e\in\mathcal H_f\right],
\label{eq:ceg}\\
\mathrm{SUE}^{a}_{k,f}&=\mathbb{E}\!\left[Y^{a}_{c,k,e,r}-Y^{a,\mathrm{off}}_{c,k,e,r}\mid e\in\mathcal H_f\right].
\label{eq:sue}
\end{align}
A positive CEG shows that the updated agent performs better than the non-evolving baseline, but it does not by itself demonstrate capability acquisition. We attribute the gain to the target capability only when it also appears on held-out episodes and remains under controlled environmental variations. A positive SUE further shows that the improvement depends on the learned artifact, because performance decreases when the artifact is unavailable. Together, these results distinguish a transferable capability from an isolated increase in terminal wealth.

\paragraph{Learning generalization.}
Within each \emph{Accumulation} stream for task family $f$, we evaluate the artifact from checkpoint $K_5$ on near-distribution $\mathcal{H}_f^{\mathrm{N}}$ and transfer $\mathcal{H}_f^{\mathrm{T}}$ held-out episodes. The near-distribution episodes $\mathcal{H}_f^{\mathrm{N}}$ vary the market realization and parameters. The transfer episodes $\mathcal{H}_f^{\mathrm{T}}$ further change the observation mapping or opponent configuration to introduce a controlled distribution shift. We report $\mathrm{CEG}_{\mathrm N}$ and $\mathrm{CEG}_{\mathrm T}$, and calculate $\mathrm{SUE}_{\mathrm H}$ accordingly. 

\paragraph{Capability retention.}
In an \emph{Interference} stream, the focal capability is established at $K_2$ and retested on the same retention episode after three distractor updates at
$K_5$. For $e\in\mathcal H_f^{\mathrm{ret}}$, we define the retention loss as $L^{a}_{c,e,r}=\max\!\left\{0,\,Y^{a}_{c,2,e,r}-Y^{a}_{c,5,e,r}\right\}$. Thus, $L^{a}_{c,e,r}$ records the performance drop from $K_2$ to $K_5$ and is zero when performance is maintained or improved. To account for retention loss that also occurs without artifact updates, we subtract the corresponding loss of the non-evolving baseline, both in expectation and in the tail.
\begin{align}
\Delta\mathrm{RL}^{a}&=\mathbb{E}[L^{a}_{c,e,r}]-\mathbb{E}[L^{\mathrm{baseline}}_{c,e,r}],
\\
\Delta\mathrm{CVaR}^{a}_{\tau}&=\mathrm{CVaR}_{\tau}(L^{a}_{c,e,r})-\mathrm{CVaR}_{\tau}(L^{\mathrm{baseline}}_{c,e,r}).
\end{align}
Here, $\mathrm{CVaR}_{\tau}$ denotes the mean of the worst $\tau$ fraction of retention losses. We report $\tau=0.2$ and $\tau=0.1$,
corresponding to the worst $20\%$ and worst $10\%$ of outcomes, respectively. Lower values indicate less capability erosion than the baseline.

\paragraph{Rule adaptation.}
A \emph{Reversal} stream changes the relation that the agent learned during its first three updates. At $K_3$, before observing any example from the new relation, we evaluate the agent on a hidden post-change episode. We evaluate the same episode again at $K_4$, after one post-change learning episode, and at $K_5$, after two. The resulting score changes show how quickly the agent replaces an outdated rule. To remove score changes that occur without artifact updates, we subtract the corresponding change of the non-evolving baseline. For the same post-change episode $e\in\mathcal H_f^{\mathrm{rev}}$, we define the revision gains as
\begin{align}
\mathrm{REV}^{a}_{1}&=\mathbb{E}\!\left[\left(Y^{a}_{c,4,e,r}-Y^{a}_{c,3,e,r}\right)-\left(Y^{\mathrm{baseline}}_{c,4,e,r}-Y^{\mathrm{baseline}}_{c,3,e,r}\right)\right],
\label{eq:revision-one}\\
\mathrm{REV}^{a}_{2}&=\mathbb{E}\!\left[\left(Y^{a}_{c,5,e,r}-Y^{a}_{c,3,e,r}\right)-\left(Y^{\mathrm{baseline}}_{c,5,e,r}-Y^{\mathrm{baseline}}_{c,3,e,r}\right)\right].
\label{eq:revision-two}
\end{align}
A positive $\mathrm{REV}^{a}_{1}$ means that method $a$ responds to the new relation after one update better than the baseline. A positive $\mathrm{REV}^{a}_{2}$ means that this adaptation is present after two updates.

\section{Experiments}
\label{sec:experiments}
This section evaluates whether self-evolution produces capabilities that generalize, survive subsequent updates, and adapt when previously useful rules become outdated. We further examine the computational costs incurred during self-evolution, as well as key design choices and their impacts. Additional experimental results can be found in Appendix~\ref{app:exp_results}.

\subsection{Experimental Setup}
\label{sec:experimental-setup}
\paragraph{Compared methods.}
We compare ten configurations in three groups. \emph{Artifact-Free Execution} comprises the baseline, which retains no cross-episode information, and Context, which provides recent interaction history without summarization. \emph{Episodic Memory} includes Reflection (storing episode-level summaries), Single-Evidence Memory (extracting rules from individual episodes), and Consolidated Memory (activating rules only after corroboration across multiple episodes). Finally, \emph{Skill Evolution} encompasses SkillOpt \cite{yang2026skilloptexecutivestrategyselfevolving}, SkillBoost \cite{lin2026rethinkingselfevolutionconstrainedexplorationexploitation}, SkillX \cite{wang2026skillx}, Trace2Skill \cite{ni2026trace2skilldistilltrajectorylocallessons}, and SkillGrad \cite{wang2026skillgradoptimizingagentskills}. All evolving methods follow a cycle: executing an episode, observing feedback, updating a persistent artifact, and invoking it subsequently. Some methods additionally perform paired internal validation rollouts before committing an update to reduce regressions on previously observed cases. Details for these methods are provided in Appendix~\ref{app:skill-evolution-methods}. 

\paragraph{Evaluation protocol.}
To ensure fair comparison, all methods share identical settings, seeds, and constraints. Each completes five learning episodes of 24 market steps and three decisions. Post-episode feedback allows the agent to update its memories or skills for future use. We evaluate each method before learning and after every update opportunity, giving six checkpoints ($K_0,\ldots,K_5$). At each checkpoint, we copy the stored memories or skills and test the copy on held-out episodes. We examine whether learning improves performance on new tasks, whether learning other tasks reduces earlier performance, and whether new evidence helps the agent revise a previously useful rule. 

\begin{table*}[t]
\centering
\caption{Main results on \textsc{EvoPathBench}. Arrows indicate the preferred direction. Bold and underline denote the best and second-best non-baseline estimates.}
\label{tab:main-results}
\vspace{3pt}
\begingroup
\setlength{\tabcolsep}{3.5pt}
\renewcommand{\arraystretch}{1.15}
\scriptsize

\resizebox{\textwidth}{!}{%
\begin{tabular}{
  @{}
  >{\centering\arraybackslash}m{2.75cm}
  *{8}{c}
  @{}
}
\toprule
& \multicolumn{3}{c}{\textbf{Learning generalization}}
& \multicolumn{3}{c}{\textbf{Capability retention}}
& \multicolumn{2}{c}{\textbf{Rule adaptation}} \\
\cmidrule(lr){2-4}
\cmidrule(lr){5-7}
\cmidrule(lr){8-9}

\textbf{Method}
& \makecell{Near\\$\mathrm{CEG}_{\mathrm N}\uparrow$}
& \makecell{Transfer\\$\mathrm{CEG}_{\mathrm T}\uparrow$}
& \makecell{State effect\\$\mathrm{SUE}_{\mathrm H}\uparrow$}
& \makecell{Mean\\$\Delta\mathrm{RL}\downarrow$}
& \makecell{Worst $20\%$\\$\Delta\mathrm{CVaR}_{0.2}\downarrow$}
& \makecell{Worst $10\%$\\$\Delta\mathrm{CVaR}_{0.1}\downarrow$}
& \makecell{After 1 update\\$\mathrm{REV}_{1}\uparrow$}
& \makecell{After 2 updates\\$\mathrm{REV}_{2}\uparrow$} \\
\midrule

\rowcolor{gray!14}
\multicolumn{9}{c}{\textbf{Artifact-Free Execution}} \\
\addlinespace[1pt]
Baseline
& 0.000 & 0.000 & -- & 0.000 & 0.000 & 0.000 & 0.000 & 0.000 \\
Context
& 0.540 & -0.061 & 0.160
& \textbf{-0.685} & \textbf{-3.229} & \textbf{-5.505}
& -0.059 & 0.021 \\
\addlinespace[2pt]

\rowcolor{gray!14}
\multicolumn{9}{c}{\textbf{Episodic Memory}} \\
\addlinespace[1pt]
Reflection
& 0.169 & 0.162 & 0.015
& -0.360 & -1.637 & -2.597
& -0.243 & -0.331 \\
Single-Evidence Memory
& -0.460 & -0.166 & -0.208
& -0.312 & -1.369 & -2.270
& 0.471 & \textbf{0.716} \\
Consolidated Memory
& -0.036 & 0.062 & 0.127
& -0.255 & -1.202 & -2.010
& -0.031 & -0.180 \\
\addlinespace[2pt]

\rowcolor{gray!14}
\multicolumn{9}{c}{\textbf{Skill Evolution}} \\
\addlinespace[1pt]
SkillOpt
& 0.763 & -0.273 & -0.032
& 0.528 & 2.767 & 5.204
& -0.584 & 0.414 \\
SkillBoost
& \textbf{1.597} & \textbf{0.537} & \textbf{1.238}
& -0.207 & -1.044 & -1.662
& \textbf{0.666} & 0.284 \\
SkillX
& 0.252 & -0.091 & 0.089
& \underline{-0.662} & \underline{-3.096} & \underline{-5.105}
& -0.060 & 0.187 \\
Trace2Skill
& \underline{0.822} & \underline{0.293} & \underline{0.450}
& -0.476 & -2.164 & -3.143
& -0.635 & 0.225 \\
SkillGrad
& 0.555 & -0.383 & 0.005
& -0.219 & -1.142 & -2.112
& \underline{0.542} & \underline{0.619} \\
\bottomrule
\end{tabular}%
}
\endgroup
\end{table*}

\begin{figure}[t!]
\centering
\includegraphics[width=0.99\textwidth]{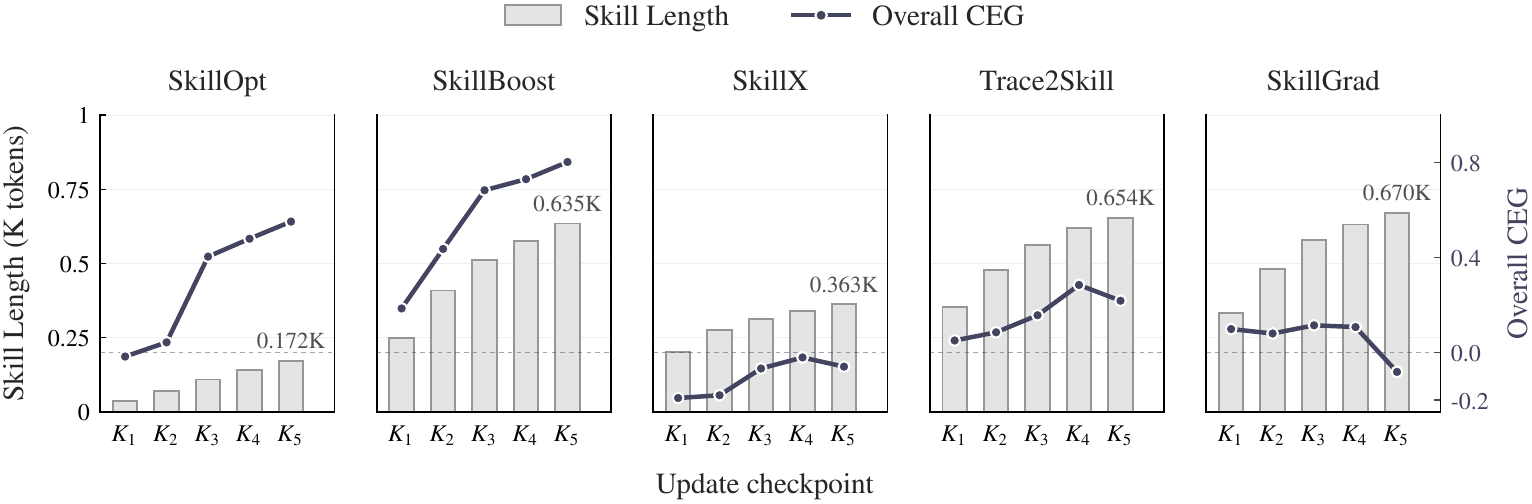}
\caption{Skill length and overall CEG across update checkpoints. Negative CEG indicates performance below the non-evolving baseline.}
 \label{fig:skill-length-ceg}
\end{figure}

\subsection{Main Results}
\paragraph{Learning generalization} Table~\ref{tab:main-results} reveals a generalization gap consistent with \textbf{overfitting}: SkillOpt and SkillGrad improve on near-distribution episodes (0.763 and 0.555) but perform worse than the baseline on transfer episodes (-0.273 and -0.383). Their updates benefit familiar conditions without generalizing to new ones. SkillBoost achieves the strongest gains on both sets (1.597 and 0.537), supported by a positive state-use effect (1.238), although its transfer benefit remains smaller.
\paragraph{Capability retention.}
SkillOpt achieves near-distribution gains at the cost of increased mean retention loss and $\mathrm{CVaR}_{0.1}$. Context and SkillX exhibit lower retention losses on both metrics yet fail to achieve transfer gains, demonstrating that limited forgetting does not guarantee successful learning. In contrast, SkillBoost attains positive generalization gains while reducing retention losses, striking a more favorable balance. These results underscore that retention should be evaluated jointly with acquisition rather than optimized in isolation.
\paragraph{Rule adaptation.}
Strong generalization does not guarantee effective revision of outdated rules. Single-Evidence Memory achieves the highest average $\mathrm{REV}_2$ despite negative generalization gains, whereas SkillBoost excels in generalization yet exhibits diminished revision benefits after its second update. This divergence reveals a fundamental challenge: acquiring useful rules does not ensure their correction upon encountering contradictory evidence. Notably, no revision effect remains statistically significant after Holm correction. Thus, the observed rankings do not constitute reliable evidence of superior adaptation. 

\begin{tcolorbox}[colback=blue!4,colframe=blue!60!white,colbacktitle=blue!60!white,coltitle=white,fonttitle=\bfseries, title={Takeaway}, boxrule=0.6pt,arc=2mm,left=2mm,right=2mm,top=1.5mm,bottom=1.5mm]
Acquiring useful rules does not guarantee the ability to correctly apply, retain, or revise them in the face of contradictory evidence. Current skill evolution methods achieve gains under familiar conditions, yet these improvements often coincide with poor transfer and capability erosion, while reliable rule revision remains unestablished.
\end{tcolorbox}

\begin{table*}[t]
\centering
\caption{Token usage and efficiency on \textsc{EvoPathBench}. Token counts are in millions. Direct update denotes the tokens consumed by model calls that create or revise the persistent artifact. Bold marks the highest CEG and TE.}
\label{tab:token-efficiency}

\begingroup
\small
\setlength{\tabcolsep}{6pt}
\renewcommand{\arraystretch}{1.12}

\begin{tabular}{@{}crrrrrr@{}}
\toprule
& \multicolumn{4}{c}{Token usage (M)}
& \multicolumn{2}{c}{Performance and efficiency} \\
\cmidrule(lr){2-5}
\cmidrule(lr){6-7}
Method
& Input
& Output
& Total
& Direct update
& CEG $\uparrow$
& TE $\uparrow$ \\
\midrule

\rowcolor{gray!12}
\multicolumn{7}{c}{\textbf{Artifact-free Execution}} \\
Baseline
& 30.96 & 2.93 & 33.88 & 0.00 & 0.000 & -- \\
Context
& 137.80 & 3.81 & 141.61 & 0.00 & -0.084 & -0.00078 \\

\addlinespace[2pt]
\rowcolor{gray!12}
\multicolumn{7}{c}{\textbf{Episodic Memory}} \\
Reflection
& 33.11 & 3.15 & 36.25 & 1.04 & -0.132 & -0.05562 \\
Single-Evidence Memory
& 33.32 & 3.14 & 36.45 & 1.05 & -0.171 & -0.06666 \\
Consolidated Memory
& 32.31 & 3.11 & 35.41 & 0.98 & -0.092 & -0.06020 \\

\addlinespace[2pt]
\rowcolor{gray!12}
\multicolumn{7}{c}{\textbf{Skill Evolution}} \\
SkillOpt
& 63.75 & 6.43 & 70.18 & 2.79
& 0.551 & \textbf{0.01519} \\
SkillBoost
& 107.35 & 13.30 & 120.65 & 21.96
& \textbf{0.802} & 0.00925 \\
SkillX
& 45.65 & 5.26 & 50.91 & 6.37
& -0.059 & -0.00347 \\
Trace2Skill
& 57.85 & 6.37 & 64.23 & 10.61
& 0.218 & 0.00719 \\
SkillGrad
& 55.78 & 6.18 & 61.95 & 8.37
& -0.081 & -0.00289 \\

\bottomrule
\end{tabular}
\endgroup
\end{table*}

\subsection{Does More Computation Lead to Better Evolution?}
\paragraph{Skill growth versus capability growth.}
We summarize performance at checkpoint $K_t$ using the overall CEG: $\mathrm{CEG}^{a}_{k,f}$ (Eq.~\ref{eq:ceg}). Figure~\ref{fig:skill-length-ceg} shows that while skill length increases across all methods, performance trajectories diverge. SkillOpt achieves sustained gains with the shortest artifacts, whereas SkillGrad, Trace2Skill and SkillX produce longer artifacts but suffer from negative CEG or diminishing returns. Only SkillBoost combines artifact growth with strong final performance, demonstrating that longer skills are beneficial but insufficient for improvement. This clearly distinguishes mere text accumulation from effective capability acquisition.

\paragraph{Token efficiency.}
We define Token Efficiency (TE) as the capability evolution gain $\mathrm{CEG}^{a}_{k,f}$ relative to the baseline: $\mathrm{TE}^{a}=\frac{\mathrm{CEG}^{a}_{k,f}}{(T^{a}-T^{\mathrm{baseline}})/10^6}$, where $T^{a}$ denotes total tokens for execution, validation, and updates. TE is defined only for positive token overhead; negative values indicate increased expenditure without commensurate gains. Table~\ref{tab:token-efficiency} shows that the strongest method is not the most token-efficient. While SkillBoost achieves the highest CEG, it requires $2.39\times$ the tokens of SkillOpt for a 45.5\% larger gain. Conversely, memory-based methods incur lower overheads but yield negative gains, and context consumes the most tokens without surpassing the baseline. 

\begin{tcolorbox}[colback=blue!4,colframe=blue!60!white,colbacktitle=blue!60!white,coltitle=white,fonttitle=\bfseries, title={Takeaway}, boxrule=0.6pt,arc=2mm,left=2mm,right=2mm,top=1.5mm,bottom=1.5mm]
More text and tokens do not yield better self-evolution. The key lies in whether the update mechanism can translate feedback into effective, yet not excessive, artifact modifications.
\end{tcolorbox}

\begin{table*}[t!]
\centering
\caption{Impact of controlled update interventions. Each row reports the second setting minus the first. Expanding candidates (1 to 4) increases proposed updates, while safeguards introduce improvement and regression gates. Metrics include: \emph{Commit rate} (fraction of saved artifacts), \emph{Best gain} (top candidate improvement), \emph{Selection gap} (difference between top candidate and saved update), \emph{Harmful} (committed regressions), and \emph{Missed} (rejected improvements)}
\label{tab:update-mechanisms}
\vspace{2pt}

\begingroup
\small
\setlength{\tabcolsep}{2.4pt}
\renewcommand{\arraystretch}{1.12}
\sisetup{
  detect-weight=true,
  retain-explicit-plus=true
}

\resizebox{\textwidth}{!}{%
\begin{tabular}{
@{}l
S[table-format=+1.3]
S[table-format=+1.3]
S[table-format=+1.3]
S[table-format=+1.3]
S[table-format=+1.3]
S[table-format=+1.3]
S[table-format=+1.3]
S[table-format=+1.3]
S[table-format=+1.3]
S[table-format=+1.3]@{}
}
\toprule[0.9pt]
& \multicolumn{5}{c}{\textbf{Update-process change}}
& \multicolumn{5}{c}{\textbf{Held-out performance change}} \\
\cmidrule(lr){2-6}
\cmidrule(lr){7-11}

\multicolumn{1}{l}{\textbf{Intervention}}
& \multicolumn{1}{c}{\shortstack{\textbf{Commit}\\\textbf{rate}}}
& \multicolumn{1}{c}{\shortstack{\textbf{Best}\\\textbf{gain} $\uparrow$}}
& \multicolumn{1}{c}{\shortstack{\textbf{Selection}\\\textbf{gap} $\downarrow$}}
& \multicolumn{1}{c}{\shortstack{\textbf{Harmful}\\\textbf{rate} $\downarrow$}}
& \multicolumn{1}{c}{\shortstack{\textbf{Missed}\\\textbf{rate} $\downarrow$}}
& \multicolumn{1}{c}{$\Delta\mathrm{CEG}_{\mathrm N}\uparrow$}
& \multicolumn{1}{c}{$\Delta\mathrm{CEG}_{\mathrm T}\uparrow$}
& \multicolumn{1}{c}{$\Delta\mathrm{RL}\downarrow$}
& \multicolumn{1}{c}{$\Delta\mathrm{REV}_{1}\uparrow$}
& \multicolumn{1}{c}{$\Delta\mathrm{REV}_{2}\uparrow$}
\\
\midrule

1 candidate $\rightarrow$ 4 candidates
& +0.180 & +0.496 & +0.163 & +0.007 & -0.029
& +0.186 & -1.463 & -0.115 & -0.064 & +0.419
\\
\addlinespace[1.5pt]

Open $\rightarrow$ safeguarded acceptance
& -0.607 & +0.690 & +0.253 & -0.033 & +0.151
& +0.368 & +1.420 & -0.300 & +0.422 & +0.121
\\

\bottomrule[0.9pt]
\end{tabular}%
}

\endgroup
\end{table*}

\begin{table*}[t!]
\centering
\caption{Effect of safeguards on candidate selection. Both policies are compared at the same checkpoints, and candidates are tested on unseen episodes that are not used for selection}
\label{tab:candidate-selection}

\vspace{2pt}
\begingroup
\small
\setlength{\tabcolsep}{3.5pt}
\renewcommand{\arraystretch}{1.15}
\sisetup{
    detect-weight=true,
    table-number-alignment=center
}

\resizebox{\textwidth}{!}{%
\begin{tabular}{
@{}
l
S[table-format=2.1]
S[table-format=2.1]
S[table-format=2.1]
S[table-format=2.1]
S[table-format=2.1]
S[table-format=2.1]
@{}
}
\toprule[0.9pt]

& \multicolumn{2}{c}{\textbf{Validation fidelity}}
& \multicolumn{2}{c}{\textbf{Utility selection}}
& \multicolumn{2}{c}{\textbf{Decision error}}
\\
\cmidrule(lr){2-3}
\cmidrule(lr){4-5}
\cmidrule(lr){6-7}

\multicolumn{1}{l}{\textbf{Acceptance}}
& \multicolumn{1}{c}{\textbf{Agreement} $(\%)\uparrow$}
& \multicolumn{1}{c}{\textbf{Capture} $(\%)\uparrow$}
& \multicolumn{1}{c}{\textbf{Precision} $(\%)\uparrow$}
& \multicolumn{1}{c}{\textbf{Recall} $(\%)\uparrow$}
& \multicolumn{1}{c}{\textbf{Harmful} $(\%)\downarrow$}
& \multicolumn{1}{c}{\textbf{Missed} $(\%)\downarrow$}
\\
\midrule

Open
& 56.9
& 69.0
& 9.3
& 52.9
& 12.4
& 0.0
\\

Safeguarded
& 55.8
& 47.2
& 29.5
& 41.9
& 6.2
& 16.5
\\

\bottomrule[0.9pt]
\end{tabular}
}
\endgroup
\end{table*}

\subsection{What Makes Self-Evolution Effective?}
\label{sec:useful-evolution}
At each update point, the agent proposes several candidate skill documents. The acceptance rule either saves one candidate or retains the current skill. After this decision, we independently evaluate every candidate and the current skill on unseen episodes that are never used for selection. This allows us to compare what the selection-time validation predicts with how each candidate actually performs on held-out tasks.

\paragraph{Generation alone is insufficient.}
We intervene on two components of a generic skill updater: candidate search and acceptance safeguards. Table~\ref{tab:update-mechanisms} reports both changes in the update process and their downstream effects. Expanding the candidate pool produces stronger candidates, but also increases the gap between the best available candidate and the update actually selected. These results identify candidate selection, rather than candidate generation, as a central bottleneck.

\paragraph{Can the acceptance rule identify useful updates?}Table~\ref{tab:candidate-selection} reports six complementary measures: \emph{Agreement} (consistency between validation and held-out evaluations), \emph{Capture} (fraction of potential gain realized), \emph{Precision} (accepted updates improving held-out performance), \emph{Recall} (selection rate of beneficial candidates), \emph{Harmful} (accepted regressions), and \emph{Missed} (unselected superior candidates). Results indicate that safeguards alter error patterns without enhancing held-out prediction. While safeguards improve precision and reduce harmful updates, they lower capture and recall, increasing missed improvements. Overall, safeguards provide greater stability mainly by accepting fewer updates, rather than by identifying better candidates.

\begin{table}[h]
\centering

\begin{minipage}[t]{0.49\textwidth}
\vspace{0pt}
\paragraph{Do evolved skills transfer across models?}
We transfer skills learned by Qwen3.8-Max to frozen receiver models and compare
them with SFT on inference trajectories from the same source model.
Table~\ref{tab:cross-model-transfer} shows that evolved skills encode
transferable knowledge and can produce gains comparable to SFT without
parameter updates. Their effectiveness depends on the receiver.
\end{minipage}
\hfill
\begin{minipage}[t]{0.48\textwidth}
\vspace{0pt}
\centering
\caption{Cross-model transfer results. Overall CEG relative to Base. Higher is better.}
\label{tab:cross-model-transfer}
\vspace{3pt}

\small
\setlength{\tabcolsep}{3.5pt}
\renewcommand{\arraystretch}{1.10}
\sisetup{
  mode=math,
  table-number-alignment=center,
  table-text-alignment=center,
  table-format=+1.2,
  detect-weight=true,
  retain-explicit-plus=true,
  round-mode=places,
  round-precision=2,
  round-pad=true
}

\begin{tabular*}{\linewidth}{
@{\extracolsep{\fill}}
l
S
S
S
@{}
}
\toprule
\textbf{Receiver}
& \multicolumn{1}{c}{\textbf{SkillBoost}}
& \multicolumn{1}{c}{\textbf{SkillOpt}}
& \multicolumn{1}{c}{\textbf{SFT}}
\\
\midrule
Qwen3.5-4B
& +0.27
& +0.19
& +0.14
\\
Qwen3.5-9B
& +0.03
& -0.02
& +0.11
\\
Qwen3.8-27B
& +1.96
& +0.46
& +2.19
\\
\bottomrule
\end{tabular*}
\end{minipage}

\end{table}

\begin{tcolorbox}[colback=blue!4,colframe=blue!60!white,colbacktitle=blue!60!white,coltitle=white,fonttitle=\bfseries, title={Takeaway}, boxrule=0.6pt,arc=2mm,left=2mm,right=2mm,top=1.5mm,bottom=1.5mm]
Current self-evolution methods encode actionable knowledge and achieve SFT-like gains without parameter updates. Further progress depends primarily on candidate evaluation rather than candidate generation. Broader search produces stronger candidates, yet current acceptance rules do not reliably distinguish transferable improvements from overfitting or harmful forgetting.\end{tcolorbox}

\subsection{How Does Forgetting Manifest in Self-Evolving Agents?}
For each retention episode $e\in\mathcal H_f^{\mathrm{ret}}$, let $D^{a}_{c,e,r}=Y^{a}_{c,5,e,r}-Y^{a}_{c,2,e,r}$ denote the signed change and $L^{a}_{c,e,r}=\max\{0,-D^{a}_{c,e,r}\}$ the corresponding loss. We characterize retention using the backward transfer $\mathrm{BWT}^{a}=\mathbb{E}[D^{a}_{c,e,r}]$, forgetting rate $P_{\mathrm F}^{a}=\Pr(L^{a}_{c,e,r}>0)$, conditional loss $M_{\mathrm F}^{a}=\mathbb{E}[L^{a}_{c,e,r}\mid L^{a}_{c,e,r}>0]$, and $\mathrm{CVaR}_{0.1}(L^a)$. The baseline captures background variation between $K_2$ and $K_5$.

\paragraph{Forgetting is sparse and tail-dominated.}
Table~\ref{tab:forgetting-decomposition} shows that several methods exhibit little forgetting. However, their limited transfer gains and small SUE values in Table~\ref{tab:main-results} suggest that they also acquire few useful capabilities. SkillOpt has a lower forgetting rate than the baseline but the largest mean loss $\mathbb{E}[L^a]$ and $\mathrm{CVaR}_{0.1}$. Its poor retention therefore reflects larger losses when forgetting occurs, rather than more frequent forgetting. SkillBoost achieves the highest positive BWT, yet performance still declines on $21.3\%$ of trajectories. Across methods, $\mathrm{CVaR}_{0.1}$ substantially exceeds mean retention loss, showing that losses are concentrated in a minority of trajectories. %These results distinguish average improvement from reliable preservation of previously acquired capabilities.

\begin{tcolorbox}[colback=blue!4,colframe=blue!60!white,colbacktitle=blue!60!white,coltitle=white,fonttitle=\bfseries, title={Takeaway}, boxrule=0.6pt,arc=2mm,left=2mm,right=2mm,top=1.5mm,bottom=1.5mm]
Several self-evolution methods perform close to the baseline, showing limited improvement alongside relatively little forgetting. Leading methods achieve larger average gains but still suffer severe losses on a minority of trajectories, exposing
a substantial tail risk. These findings motivate update mechanisms that identify which rules require revision and which capabilities should be preserved.
\end{tcolorbox}

\begin{table*}[t!]
\centering
\caption{Decomposition of retention outcomes from $K_2$ to $K_5$ in \emph{Interference} streams. Forgetting rate measures how often performance declines, conditional loss measures the average decline when forgetting occurs, and $\mathrm{CVaR}_{0.1}$ is the mean retention loss among the worst $10\%$ of paths. Bold indicates the best point estimate in each column.}
\label{tab:forgetting-decomposition}

\vspace{2pt}
\begingroup
\footnotesize
\setlength{\tabcolsep}{1.5pt}
\renewcommand{\arraystretch}{1.05}
\sisetup{
  detect-weight=true,
  table-number-alignment=center,
  retain-explicit-plus=true,
  round-mode=places,
  round-precision=3,
  round-pad=true
}

\resizebox{\textwidth}{!}{%
\begin{tabular}{
@{}
l
*{10}{S[table-format=+2.3]}
@{}
}
\toprule[0.9pt]
& \multicolumn{2}{c}{\scriptsize\textbf{Artifact-free}}
& \multicolumn{3}{c}{\scriptsize\textbf{Memory}}
& \multicolumn{5}{c}{\scriptsize\textbf{Skill evolution}}
\\
\cmidrule(lr){2-3}
\cmidrule(lr){4-6}
\cmidrule(lr){7-11}
\multicolumn{1}{l}{\scriptsize\textbf{Metric}}
& \multicolumn{1}{c}{\scriptsize\textbf{Baseline}}
& \multicolumn{1}{c}{\scriptsize\textbf{Context}}
& \multicolumn{1}{c}{\scriptsize\textbf{Reflection}}
& \multicolumn{1}{c}{\scriptsize\textbf{Single-Evidence}}
& \multicolumn{1}{c}{\scriptsize\textbf{Consolidated}}
& \multicolumn{1}{c}{\scriptsize\textbf{SkillOpt}}
& \multicolumn{1}{c}{\scriptsize\textbf{SkillBoost}}
& \multicolumn{1}{c}{\scriptsize\textbf{SkillX}}
& \multicolumn{1}{c}{\scriptsize\textbf{Trace2Skill}}
& \multicolumn{1}{c}{\scriptsize\textbf{SkillGrad}}
\\
\midrule

\textbf{BWT} $\uparrow$
& -0.082
& +0.319
& +0.043
& -0.090
& +0.185
& -0.721
& \bfseries +0.863
& +0.231
& +0.327
& +0.235
\\

\textbf{Mean loss} $\mathbb{E}[L^a]\downarrow$
& 0.926
& \bfseries 0.241
& 0.567
& 0.614
& 0.671
& 1.454
& 0.719
& 0.264
& 0.451
& 0.707
\\

\textbf{Forgetting rate} $(\%)\downarrow$
& 23.6
& 14.8
& 15.7
& 15.7
& 16.7
& 20.8
& 21.3
& \bfseries 10.6
& 13.4
& 24.1
\\

\textbf{Conditional loss} $M_{\mathrm F}\downarrow$
& 3.113
& \bfseries 1.495
& 2.054
& 3.091
& 2.795
& 5.631
& 3.414
& 3.221
& 3.316
& 2.469
\\

\textbf{Worst 10\%} $\mathrm{CVaR}_{0.1}\downarrow$
& 7.461
& \bfseries 1.956
& 4.864
& 5.191
& 5.452
& 12.666
& 5.799
& 2.356
& 4.319
& 5.350
\\

\bottomrule[0.9pt]
\end{tabular}%
}
\endgroup
\end{table*}

\section{Conclusion}
\label{sec:conclusion}

We present \textsc{EvoPathBench} to track the dynamics of capabilities derived from artifact-level self-evolution under subsequent updates. The benchmark uses ordered task streams and read-only checkpoint evaluations to measure learning generalization, capability retention, and rule adaptation separately. Its hybrid market environment combines public trading data with generated trajectories calibrated to similar market properties. Across the evaluated methods, gains under familiar conditions often weaken under distribution shift, retention losses concentrate in a minority of paths, and reliable rule adaptation remains unestablished. Additional search, longer artifacts, and greater token use do not consistently improve these outcomes. Our mechanism analyses identify candidate selection as a central challenge, since current acceptance rules have limited capacity to distinguish transferable improvements from overfitting, or necessary rule revision from harmful forgetting. Reliable self-evolution therefore depends on selecting artifact updates that improve new behavior while preserving capabilities that remain valid.

\subsection*{AI use statement}
In this work, we used generative AI tools to polish sentence-level writing, improve clarity and grammar, and identify potential typographical or consistency errors. We also used generative AI to draft and revise portions of the implementation and analysis code. All AI-assisted text was reviewed and edited by the authors, and all AI-generated code was manually inspected and tested against the expected behavior and experimental outputs. Generative AI was not used to generate the benchmark data; the semi-synthetic data were produced by a seeded and reproducible procedural simulator. We take full responsibility for the final content of this work, including all text, code, results, claims, and artifacts produced with the aid of generative AI.

\subsection*{Ethics statement}

This work involves no human subjects, personal data. We do not identify any specific ethical concerns arising from this research.

\bibliographystyle{plainnat}
\setcitestyle{numbers}
\bibliography{iclr2027_conference}

\clearpage
\appendix

\noindent{\Large\bfseries Appendix}
\par\vspace{0.8em}

\noindent{\large\bfseries Table of Contents}
\par\vspace{0.6em}

\startcontents[appendices]

\begingroup
% Main appendix entries.
\titlecontents{section}
  [0em]
  {\addvspace{0.55em}\bfseries}
  {\contentslabel{1.8em}}
  {}
  {\hfill\bfseries\contentspage}

% Subsection entries.
\titlecontents{subsection}
  [1.8em]
  {\addvspace{0.08em}\normalfont}
  {\contentslabel{2.6em}}
  {}
  {\titlerule*[0.7em]{.}\contentspage}

\printcontents[appendices]{}{1}{
  \setcounter{tocdepth}{2}
}
\endgroup

\clearpage

\section{Benchmark Structure}
\label{app:benchmark-composition}

\begin{figure}[t!]
\centering
\includegraphics[scale=0.6]{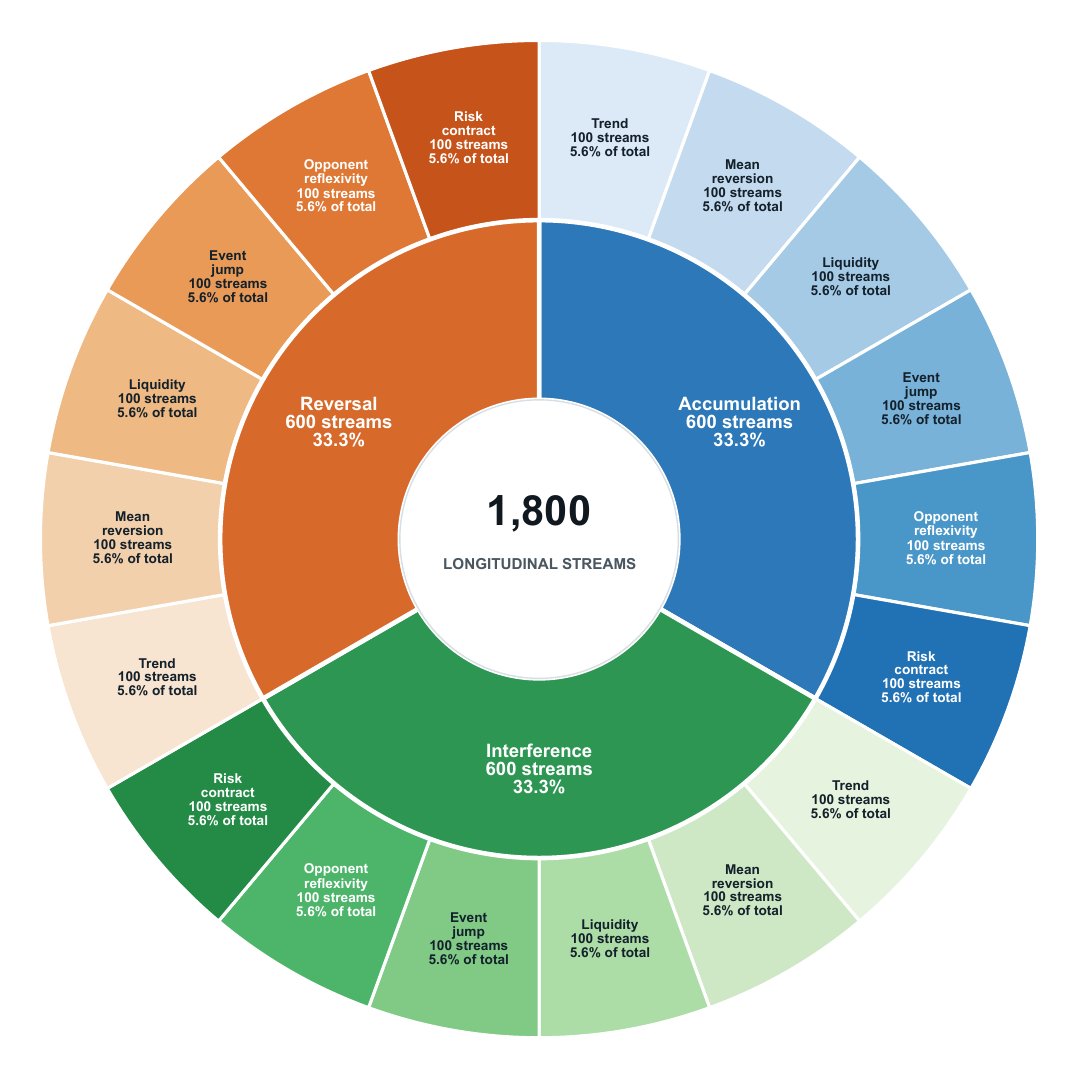}
\caption{Composition of the EvoPathBench benchmark. The inner ring represents the three stream templates, and the outer ring represents the six focal task families. Sector area is proportional to the number of streams. The benchmark contains 600 streams per template and 100 streams in each template--family cell. Each cell includes streams from both market layers.}
 \label{fig:benchmark-composition}
\end{figure}

Figure~\ref{fig:benchmark-composition} summarizes the composition of the EvoPathBench. The benchmark combines three longitudinal stream templates with six controlled task families. Each stream contains five episodes.

\subsection{Longitudinal Streams}
The three stream templates isolate different aspects of capability acquisition, retention, and revision. Each stream contains five sequential episodes, and model updates are permitted after every episode.

\emph{Accumulation} presents five episodes from the same focal task family. Repeated exposure allows evidence about a common market mechanism to accumulate over time. This template evaluates whether an agent can identify the recurring
mechanism, distinguish persistent evidence from episode-specific noise, compress experience into a reusable representation, and apply the acquired capability in later episodes. It also reveals whether learning improves progressively or remains dependent on the most recent observation.

\emph{Interference} begins with two episodes from the focal task family. These are followed by three episodes drawn from distinct distractor families. The distractor episodes introduce unrelated learning signals without directly repeating the focal mechanism. This template evaluates whether the capability acquired during the first two episodes remains accessible after subsequent updates. It also tests whether the agent preserves the scope of that capability and avoids applying it indiscriminately to unrelated market conditions.

\emph{Reversal} begins with three episodes governed by the original relation. The final two episodes provide systematic counterevidence generated under the opposite relation. This template evaluates whether the agent detects that its previous rule is no longer valid, revises the rule in response to consistent counterevidence, and retains the revised capability. It further measures whether the agent adapts cautiously when the available evidence conflicts with previously consolidated knowledge.

\subsection{Task Families}
The six task families introduce targeted changes to the calibrated market process. Each family isolates a market mechanism and defines the capability that an agent must acquire.

\begin{itemize}
\item \textbf{Trend.}
This family introduces a persistent directional component into price dynamics. The resulting price path contains a sustained tendency to move upward or downward, although short-term fluctuations remain present. The intervention tests whether an agent can identify directional persistence, separate it from local noise, and adjust its exposure consistently with the inferred direction. It also tests whether the agent can avoid overreacting to temporary movements against the prevailing trend.

\item \textbf{Mean reversion.}
This family creates a temporary deviation between the market price and its underlying fundamental value. A corrective force causes the deviation to decrease over time. The intervention tests whether an agent can distinguish a temporary displacement from a persistent change in value, estimate the direction of correction, and time its entry and exit accordingly. Successful adaptation also requires the agent to reduce its position as the deviation closes.

\item \textbf{Liquidity.}
This family increases the cost and uncertainty of trade execution. It combines wider bid-ask spreads, lower market depth, higher transaction fees, and stronger microstructure noise. These changes make aggressive or frequent trading less effective even when the agent predicts the direction of price movement correctly. The intervention tests whether an agent adapts its order size, execution timing, and turnover to prevailing liquidity conditions. It also measures whether the agent accounts for trading costs when evaluating the profitability of a strategy.

\item \textbf{Event jump.}
This family increases the frequency and magnitude of abrupt price changes. Shock sizes follow a heavy-tailed pattern, and individual shocks may have temporary or permanent effects. The intervention tests whether an agent can recognize discontinuous changes, update its beliefs after a large shock, and distinguish persistent repricing from temporary disruption. It also evaluates whether the agent limits exposure when jump risk makes recent price behavior an unreliable guide to near-term outcomes.

\item \textbf{Opponent reflexivity.}
This family strengthens the dependence of future prices on recent order flow and on the actions of other market participants. Price dynamics therefore become more endogenous, and the effect of an action depends partly on the behavior that it induces in other agents. The intervention tests whether an agent can recognize strategic feedback, account for the market impact of recent actions, and revise its policy when previously profitable behavior changes the subsequent environment.

\item \textbf{Risk contract.}
This family combines volatile and jump-prone market conditions with binding constraints on position size, individual order size, turnover, drawdown, and short selling. The agent must therefore optimize performance within an explicitly restricted feasible set. The intervention tests whether the agent can control exposure, preserve sufficient capacity for future actions, and respond to adverse conditions without violating the contract. It also distinguishes genuine risk-aware adaptation from strategies that obtain high returns by taking prohibited or unsustainable risks.

\end{itemize}

These task families are diagnostic interventions defined relative to the calibrated market process. They are designed to isolate specific learning and adaptation requirements. They do not constitute a natural, complete, or mutually exclusive taxonomy of financial markets.

\subsection{Market Layer}
Each task family is instantiated under one of two market layers, which differ in whether price formation responds to the focal agent's own trading. This distinction lets us separate an agent's ability to exploit a fixed environment from its ability to act within one that reacts to its behavior.
\begin{itemize}
\item \textbf{Exogenous layer.}
This layer fixes the market path so that price dynamics evolve independently of the focal agent's actions. Prices are driven entirely by exogenous factors, such as latent fundamentals, volatility, and stochastic shocks. Consequently, the agent's orders affect only its own holdings and cash balance without impacting the quoted price. This setup isolates a pure forecasting-and-timing setting where the environment is a given trajectory. It tests whether an agent can infer the properties of an exogenous process, determine the size and timing of its exposure to the anticipated path, and improve its decisions purely from observation, free from confounding by its own market footprint.

\item \textbf{Endogenous layer.}
In this layer, the focal agent trades in a shared call auction against five fixed opponents following distinct behavioral rules, spanning fundamental valuation, momentum, contrarian, market-making, and noise trading. At each step, the orders from the focal agent and its opponents are matched, and the resulting trades determine the execution price rather than a pre-set path. The net buying or selling pressure across all participants creates an order imbalance, which feeds into a persistent, decaying price-impact term. Consequently, trades not only move the current price but also propagate into subsequent prices, while a fundamental-pull force draws the price back toward its underlying value. Price formation is thus endogenous and reflexive: the effect of an action depends partly on how it shifts the market and on the responses it induces in others. This layer tests whether an agent can account for the immediate and lagged market impact of its own trades, manage execution against strategically responsive counterparties, and revise a previously profitable policy once its actions alter the subsequent price path. It also distinguishes genuine impact-aware adaptation from strategies that appear profitable only when market feedback is ignored.
\end{itemize}

\subsection{Scope and Cross-Domain Applicability}
\label{app:cross-domain-scope}

EvoPathBench evaluates self-evolution along three core dimensions: the generalization of learned capabilities to unseen tasks, the \textit{persistence} of these capabilities after exposure to unrelated tasks, and the \textit{adaptability} to revise rules when new evidence invalidates previously useful ones. These dimensions are defined by the organization of experience and evaluation episodes rather than by domain-specific concepts. Specifically, \textit{Accumulation} streams introduce consistent evidence, \textit{Interference} streams insert unrelated learning between two evaluations of the same capability, and \textit{Reversal} streams alter the underlying relations the agent must follow. Consequently, the benchmark measures capability trajectories across controlled updates, moving beyond single-task performance or static endpoints as evidence of successful self-evolution.

In this work, we instantiate these dimensions within a trading environment, which is chosen for its support of controlled rule changes, reproducible trajectories, and quantitative evaluation of sequential decisions. Furthermore, financial market data is rarely included in the pre-training corpora of large language models, which helps mitigate the risk of benchmark leakage. The evaluation framework is transferable. It can be adapted to new domains by constructing domain-specific held-out tasks, distractor tasks, and rule reversals while retaining the same checkpoint-based comparison protocol. Extending this protocol beyond trading is therefore a critical step for validating the cross-domain robustness of our empirical conclusions.

\section{Experimental Setup}
\label{app:exp_setup}
\subsection{Episodic-Memory Methods}
\label{app:episodic-memory-methods}

We compare three methods that retain information from previous episodes without modifying the agent's model parameters. The methods differ primarily in whether this lesson is stored as an individual experience or incorporated into a reusable rule.

\paragraph{Reflection.}
Reflection stores episode-specific lessons without combining evidence across episodes. Each record contains the observable conditions, a recommended behavior, an explanation of the outcome, and the episode score. Any valid proposed lesson is appended directly, without requiring repeated evidence or demonstrated improvement on validation episodes. The memory retains a bounded number of recent records. During a subsequent episode, the agent retrieves the highest-scoring record whose observable conditions exactly match the current state, and the recommended behavior from that record is injected into the decision prompt. An individual experience can therefore guide later decisions without first being consolidated into a reusable rule.

\paragraph{Single-Evidence Memory.}
Single-evidence memory stores structured rules rather than separate episode records. Each rule specifies observable conditions, a strategy label, and a natural-language behavioral instruction, together with supporting episode identifiers, scores, and a confidence value. A single proposed lesson is sufficient to create or revise a rule. When a proposal matches an existing rule's conditions and strategy label, its evidence is added and its instruction is refreshed. A different strategy label under the same conditions replaces the previous rule. The model may also explicitly invalidate a visible rule. During execution, retrieval first seeks an exact condition match and, if none exists, considers rules sharing the same observable market regime. The highest-ranked matching rule is supplied to the agent.

\paragraph{Consolidated Memory.}
Consolidated Memory uses the same rule representation and retrieval procedure but requires two episode-level proposals with the same observable conditions and strategy label before creating or revising a rule. A proposal supported only once remains pending and is not available for execution. Once the evidence requirement is met, the
rule is added, refreshed, or replaced using the same procedure as single-evidence memory. This criterion assesses whether a proposed strategy recurs across experiences, rather than requiring identical wording or consistently positive performance. However, explicit invalidation of an observable rule bypasses this two-episode threshold.

\paragraph{Memory use and update criteria.}
Both rule-based methods track the outcomes of rules that influence decision-making. Negative outcomes reduce a rule's confidence, thereby lowering its retrieval priority and increasing its likelihood of removal when memory capacity is reached. Retrieved instructions guide the frozen agent via the prompt but do not override the standard risk protocol. Since none of these three methods employs candidate rollout validation for rule acceptance, their primary distinction lies in how experience is stored and the amount of evidence required before a proposed rule becomes active.

\subsection{Skill-Evolution Methods}
\label{app:skill-evolution-methods}
We compare five representative skill-evolution methods. Each method keeps the agent model fixed and updates an external artifact using evidence from previous task executions. Each update receives the observable execution trace, recent decision summaries, and episode outcomes, including the score, transaction costs, and constraint violations. At held-out checkpoints, the artifact is frozen. State-off evaluation removes the skill document from the prompt without modifying the saved state.

\paragraph{SkillOpt.}
SkillOpt updates a single natural-language skill document through bounded text edits \cite{yang2026skilloptexecutivestrategyselfevolving}. Given the current document and episode evidence, an optimizer proposes additions, replacements, or deletions. Edits targeting existing text must match a unique passage before they can be applied. The resulting candidate and the current document are evaluated on the same internal validation cases. The candidate is saved only if its mean score is strictly higher. Otherwise, the current document is retained. 

\paragraph{SkillBoost.}
SkillBoost maintains a skill document and searches over alternative revisions \cite{lin2026rethinkingselfevolutionconstrainedexplorationexploitation}. A diagnosis first identifies weaknesses in the observed execution and behaviors that should be preserved. This diagnosis guides several candidate documents emphasizing conservative repair, the observed failure, risk control, or transferability. Each candidate is evaluated against the current document on matched internal validation cases. A candidate is eligible only if it
improves the mean score and satisfies the configured limits on case-level and validation-group regressions. The eligible candidate with the largest mean improvement is saved. If none qualifies, the document remains unchanged.

\paragraph{SkillX.}
SkillX stores a structured library with planning, functional, and atomic skills \cite{wang2026skillx}. These levels describe strategic objectives, reusable procedures, and individual operations, respectively. Each entry contains guidance and observable conditions for applying it. An extraction step proposes additions or revisions from the latest episode. A consolidation step then merges overlapping entries and filters advice specific to an individual episode. The library is saved when the resulting structure is valid, nonempty, and changed. Acceptance relies on these checks and the consolidation output, without an additional performance-validation gate. The library is rendered as text for subsequent decisions.

\paragraph{Trace2Skill.}
Trace2Skill maintains a skill document that is revised through independently proposed patches \cite{ni2026trace2skilldistilltrajectorylocallessons}. 
At each update, multiple analyst calls receive the same episode evidence and the same frozen document. Each call proposes local changes and identifies guidance to preserve.  A consolidation call then combines the proposals, resolves conflicts, and produces a single updated document.  The document is saved if it is nonempty and differs from the current version, without a separate performance-validation gate.  This design consolidates multiple analyses of the current episode at each update opportunity.
\paragraph{SkillGrad.}
SkillGrad stores a package containing activation guidance, general instructions, and reference procedures, together with a persistent record of recurring success and failure patterns \cite{wang2026skillgradoptimizingagentskills}. A diagnosis identifies a weakness in the latest execution. A momentum step combines this diagnosis with previously recorded patterns and specifies a proposed correction. A patching step then revises the relevant part of the package. The updated state is saved when it passes structural checks and differs from the current state, without an additional performance-validation gate. The activation guidance, instructions, and references are rendered together for execution. The pattern history informs later updates.

\paragraph{Internal validation.}
For SkillOpt and SkillBoost, internal validation compares candidate documents with the current document using paired executions on cases derived from observed training experience. These cases are separate from the held-out episodes used to report benchmark performance. The remaining methods use their respective consolidation or patching procedures to decide which content to save. Thus, performance-based acceptance is a method-specific component rather than a shared requirement.

\paragraph{Update procedure.}
The following pseudocode summarizes the implementations. Auxiliary state includes the SkillX library and SkillGrad's persistent pattern history.

\begin{quote}
\small
\begin{tabbing}
\hspace{1em}\=\hspace{1em}\=\kill
\textbf{Input:} current artifact, auxiliary state, observed episode evidence\\
\textbf{SkillOpt:}\\
\> Propose bounded edits and construct one valid candidate.\\
\> Save it if its paired validation mean exceeds the current score.\\
\textbf{SkillBoost:}\\
\> Diagnose the episode and generate alternative skill documents.\\
\> Filter candidates by mean improvement and regression limits.\\
\> Save the best eligible candidate, or retain the current document.\\
\textbf{SkillX:}\\
\> Extract skill proposals and consolidate the three-level library.\\
\> Save the library if it is valid, nonempty, and changed.\\
\textbf{Trace2Skill:}\\
\> Independently propose patches against the frozen document.\\
\> Consolidate patches and save a nonempty, changed document.\\
\textbf{SkillGrad:}\\
\> Diagnose the episode and update persistent patterns.\\
\> Patch the skill package and save a valid, changed state.\\
\textbf{Execution:} render the saved artifact into the decision prompt.
\end{tabbing}
\end{quote}
\subsection{Learning and Evaluation Episode Roles}
\label{app:episode-roles}

EvoPathBench distinguishes between \emph{learning episodes} (update-enabled) and \emph{evaluation episodes} (held-out). Learning episodes occur in the main task stream, provide outcome feedback, and permit artifact updates. Evaluation episodes are executed with frozen artifacts; their outcomes are withheld from the agent and do not influence subsequent updates.

\paragraph{Near-distribution and transfer evaluation.}
Both episode types assess generalization on unseen examples from the focal task family. \emph{Near-distribution} episodes retain the observation format and opponent configuration of learning episodes while varying market paths and parameters. \emph{Transfer} episodes preserve the core decision problem but alter the observation mapping or opponent configuration, thereby testing generalization under larger yet controlled distributional shifts.

\paragraph{Interference stream for retention evaluation.}
The Interference stream begins with two learning episodes from the focal family, followed by three update-enabled \emph{distractor episodes} drawn from unrelated task families. The agent receives feedback from distractors and may update artifacts normally. A single held-out retention episode from the focal family is evaluated both before and after these updates, enabling measurement of whether prior capabilities are preserved or degraded.

\paragraph{Reversal stream for post-change evaluation.}
The Reversal stream presents three learning episodes under the original relation, then two update-enabled \emph{counterevidence episodes} governed by a reversed relation. A \emph{post-change episode} (held-out, from the focal family, following the new relation) is evaluated at three checkpoints: $K_3$ (before counterevidence), $K_4$ (after one update), and $K_5$ (after two updates). These comparisons quantify whether the agent replaces outdated rules as contradictory evidence accumulates.

\subsection{Models and Execution Interface}
\label{app:implementation-settings}

\paragraph{Base models.}
We evaluate all methods with two API-served backbones, \texttt{Qwen3.8-Max} and \texttt{Kimi-K3}. Within each backbone experiment, the same model performs trading decisions and method-specific memory or skill updates. Model parameters remain fixed throughout the experience stream.

\paragraph{Execution interface.}
All methods interact with the same simulated trading environment through structured order submission. Each episode contains 24 market steps, with model queries at steps 0, 8, and 16. A response may contain up to four orders, each specifying a buy or sell direction, a positive integer quantity, and an optional limit price. An empty order list represents no trade. The environment checks submitted orders and applies its execution and accounting rules. Malformed decision outputs produce no orders and are recorded as errors. Agents receive no browser, shell, or unrestricted external-tool access.

\subsection{Episode Score and Task Mechanisms}
\label{app:episode-score-mechanisms}

\paragraph{Episode-level score.}
We use a common risk-adjusted trading score for all held-out evaluations. Let $Y^{a}_{c,k,e,r}$ denote the score of method $a$ in evolution path $c$, using the artifact state at checkpoint $k$, on held-out episode $e$ and execution repeat $r$. Let $W_0$ and $W_T$ denote the initial and terminal liquidated wealth, $D$ the maximum drawdown, $\bar D$ the permitted drawdown, and $N_{\mathrm{vio}}$ the number of distinct constraint violations. The score is
\begin{equation}
Y^{a}_{c,k,e,r}
=
\frac{W_T}{W_0}-1
-0.01N_{\mathrm{vio}}
-2\max\!\left\{0,D-\bar D\right\}.
\label{eq:appendix-episode-score}
\end{equation}
Terminal wealth is computed after liquidating the remaining position and therefore reflects transaction fees, bid--ask spread, and market impact. Violations include invalid orders and breaches of the position, order-size, short-selling, or turnover limits. Thus, $Y$ measures realized trading utility subject to the episode's risk contract.

\paragraph{How task mechanisms enter the evaluation.}
The scoring function does not include family-specific rewards. Instead, each task family modifies the market mechanism used to obtain the common score. Trend and mean-reversion tasks change the temporal relationship between price and fundamental value. This requires the agent to distinguish directional persistence from convergence. Liquidity tasks vary the spread and market depth, which makes trade timing and order size critical. Event-jump tasks introduce random and heavy-tailed shocks that can be either permanent or transitory. In opponent-reflexivity tasks, order flow influences subsequent prices. Therefore, the consequences of an action extend beyond its immediate execution. Risk-contract tasks impose stricter trading constraints and thus reward decisions that are compatible with these limits. Ultimately, these mechanisms require different responses without directly encoding the intended behavior in the reward.

\paragraph{From episode score to capability measurement.}
A single value of $Y$ shows how well the agent performed in one market realization. However, it does not reveal the specific strategy or capability behind this outcome. Therefore, we infer capabilities by comparing matched episodes that are held out. Near-distribution and transfer episodes test whether the learned performance can generalize to unseen situations. Retention episodes compare the same capability before and after distractor updates. Reversal episodes evaluate performance before and after the agent receives counterevidence.  Finally, a series of metrics reflecting the capabilities learned by the agent are calculated based on these controlled differences in $Y$ instead of isolated episode scores.

\subsection{Environment Representation and Agent Interaction}
\label{app:environment-interface}

The benchmark is distributed as an executable environment.  Its \texttt{episodes.jsonl} file uses JSON Lines format: each line is one complete JSON object describing an episode.  The record specifies how the simulator should construct the episode; it does not contain a precomputed sequence of 24 prices.  The simulator generates that sequence at runtime from the recorded parameters and random seed.  The separate \texttt{streams.jsonl} file lists the episodes encountered during learning and the held-out episodes evaluated at checkpoints $K_0,\ldots,K_5$.

\paragraph{Example record.}
Listing~\ref{lst:episode-json} presents a shortened version of an actual test record extracted from \texttt{episodes.jsonl}. The displayed object represents valid JSON, with all values faithfully reproduced from the released dataset. For clarity, we have omitted low-level calibration coefficients that are not essential for understanding the interface.

\begin{lstlisting}[
basicstyle=\ttfamily\scriptsize,
breaklines=true,
frame=single,
columns=fullflexible,
caption={A compact view of one episode specification.},
label={lst:episode-json}
]
{
  "episode_id": "bcal-mean_reversion-exogenous-probe_transfer-000-6be9f690d9",
  "family_id": "mean_reversion",
  "role": "probe_transfer",
  "split": "test",
  "hidden": true,
  "layer": "exogenous",
  "horizon": 24,
  "initial_price": 105.6413,
  "environment_seed": 729502753,
  "market_source": {
    "kind": "calibrated_procedural",
    "calibration_id": "bcal-4acc4a84060e4bd92f88b35a"
  },
  "mechanism_params": {
    "initial_gap": -0.025,
    "mispricing_persistence": 0.55,
    "price_volatility": 0.00502825,
    "spread_bps": 12.0,
    "liquidity": 17.0,
    "fee_bps": 2.0
  },
  "risk_contract": {
    "initial_cash": 10000.0,
    "initial_position": 20,
    "max_abs_position": 50,
    "max_order_size": 8,
    "max_turnover": 4.0,
    "max_drawdown": 0.20,
    "allow_short": false
  },
  "scoring_spec": {
    "violation_penalty": 0.01,
    "drawdown_excess_penalty": 2.0
  }
}
\end{lstlisting}

The \texttt{episode\_id} serves as the unique identifier for the record. The associated metadata---including family, role, split, and the \texttt{hidden} flag---designates this episode as a held-out mean-reversion transfer test. This setup evaluates whether a policy learned from other trajectories generalizes to a new trajectory generated by the same mechanism. Consistent with the evaluation protocol, the outcome of this episode is excluded from the evolutionary process. The environment operates under an exogenous layer assumption, meaning that while the agent's orders affect its own account, they do not influence the future market path. Note that these structural labels are utilized by the evaluator and remain hidden from the agent.The subsequent parameters define the market configuration. The episode spans 24 simulator steps, with \texttt{initial\_price=105.6413} serving as the reference price for valuing the agent's initial holdings. Prices are denominated in abstract currency units rather than US dollars. The parameter \texttt{initial\_gap=-0.025} initializes the traded price at approximately $2.5\%$ below its latent fundamental value, while \texttt{mispricing\_persistence=0.55} governs the decay rate of this price--value discrepancy. Per-step price volatility is set to approximately $0.50\%$. For context, a basis point (bp) represents one-hundredth of one percent; thus, a spread of $12$~bp corresponds to $0.12\%$, and a fee of $2$~bp equates to $0.02\%$ of the traded value. Liquidity is quantified in asset units; consequently, orders that are large relative to the baseline value of $17$ incur a more significant price-impact adjustment. Finally, the calibration identifier fixes the empirical calibration profile, and the environment seed ensures the reproducibility of the stochastic trajectory. The \texttt{risk\_contract} defines the feasible action space. The agent begins with an initial endowment of $10{,}000$ currency units and $20$ asset units. Each order is capped at $8$ units, and the position must remain within the $[0, 50]$ range as short selling is prohibited. A turnover limit of $4.0$ permits a cumulative traded value of up to four times the initial account value, while the maximum allowable drawdown is set at $20\%$. Finally, the \texttt{scoring\_spec} imposes a penalty of $0.01$ (equivalent to one percentage point) for each distinct contract violation. Additionally, it applies a penalty multiplier of two to any drawdown exceeding the permitted limit. For example, a $25\%$ drawdown under a $20\%$ threshold incurs a $10$-percentage-point penalty.

\paragraph{Information available to the agent.}
Although the stored record retains the complete data required to construct and evaluate the episode, the agent receives only a causal subset of observations. At decision steps $0$, $8$, and $16$, the agent observes historical prices and public value estimates, its current account state, execution costs, and the risk contract. Crucially, it is blinded to \texttt{family\_id}, \texttt{role}, \texttt{split}, random seeds, latent mechanism parameters, and future prices. Listing~\ref{lst:agent-observation} details the primary fields supplied at step $16$. For brevity, the preceding histories of prices and fundamentals are omitted. Here, \texttt{price} denotes the current tradable price. The public fundamental and signal serve as contemporaneous reference values; they provide no information regarding future prices or the underlying task label. The field \texttt{position=20} indicates a holding of 20 asset units. Initial wealth comprises $10{,}000$ in cash plus the initial valuation of these units, whereas marked wealth is calculated using the current price. A drawdown of $0.0042$ signifies that wealth has declined by at most $0.42\%$ from its previous peak. As the agent has not yet executed any trades, its cumulative turnover remains zero.

\begin{lstlisting}[
basicstyle=\ttfamily\scriptsize,
breaklines=true,
frame=single,
columns=fullflexible,
caption={The principal fields observed by the agent at step 16.},
label={lst:agent-observation}
]
{
  "step": 16,
  "horizon": 24,
  "price": 106.4283,
  "public_fundamental": 105.0345,
  "public_signal": 105.4181,
  "spread_bps": 12.0,
  "liquidity": 17.0,
  "account": {
    "cash": 10000.0,
    "position": 20,
    "marked_wealth": 12128.5668,
    "initial_wealth": 12112.8260,
    "max_drawdown_so_far": 0.0042,
    "turnover_so_far": 0.0
  },
  "risk_contract": {
    "max_abs_position": 50,
    "max_order_size": 8,
    "max_turnover": 4.0,
    "max_drawdown": 0.20,
    "allow_short": false
  }
}
\end{lstlisting}

\paragraph{Action and order processing.}
The current price stands $1.33\%$ above the public fundamental and $0.96\%$ above the public signal. In this instance, the agent perceives the asset as overvalued and consequently reduces its existing long position by submitting the following structured order. For readability, we have abbreviated the free-text summary while preserving the integrity of the order fields:

\begin{lstlisting}[
basicstyle=\ttfamily\scriptsize,
breaklines=true,
frame=single,
columns=fullflexible
]
{
  "orders": [{
    "side": "sell",
    "quantity": 4,
    "limit_price": 106.30,
    "tag": "mean_reversion_sell"
  }],
  "decision_summary":
    "The price exceeds both public value estimates; reduce long exposure.",
  "used_rule_ids": []
}
\end{lstlisting}

This action instructs the simulator to sell four units, subject to a limit price of at least $106.30$ per unit. The order passes all contract checks: the quantity of four is within the eight-unit limit, and the sale reduces the position from 20 to 16 without violating short-selling constraints. The traded value is approximately $425.71$, representing $3.52\%$ of the initial account value, which remains within the turnover limit. The empty \texttt{used\_rule\_ids} list indicates that no stored rule influenced this decision. This result is expected for the artifact-free baseline presented here; methods utilizing persistent memory would typically list the identifiers of the rules that affected the action. Next, the simulator accounts for execution costs. Selling incurs half of the $0.12\%$ spread, and selling four units into a liquidity depth of 17 results in a price impact of approximately $0.0235\%$. These deductions yield an executable price of $106.3394$. Since this price remains above the agent's limit of $106.30$, the order is filled. After deducting the $0.02\%$ transaction fee, the cash balance becomes $10{,}425.2727$, and the position updates to 16 units. At the subsequent step, the market price is $104.3195$, and the marked-to-market wealth is $12{,}094.3844$. The lower price results from the seeded exogenous market process and is independent of the sale; however, the account correctly reflects the completed trade. In an endogenous episode, the same order would also influence future prices through order-flow impact. The episode concludes with a return and score of $-0.1049\%$ and no constraint violations. This example illustrates the complete interaction cycle: a JSON record configures the environment, the simulator exposes current information, the agent submits a structured order, and the simulator validates, executes, and scores the resulting trajectory.

\section{Analysis and Discussion of Experimental Results}
\label{app:exp_results}
\subsection{When Historical Evaluation Conflicts with Adaptation}
\label{app:revision-case}
Rule adaptation presents a stability--plasticity trade-off. An evolving agent must revise rules invalidated by counterevidence while preserving capabilities that remain valid. Aggregate revision scores reveal whether performance eventually recovers, but they do not show when the recovery occurs or why an apparently useful update is rejected. We therefore examine adaptation at two complementary levels: checkpoint-level trajectories across methods and an internal SkillBoost validation case.

\begin{table*}[t]
\centering
\caption{Checkpoint-level performance on held-out post-reversal episodes. $K_3$ precedes counterevidence, whereas $K_4$ and $K_5$ follow the first and second update opportunities. The final column reports the observed marginal checkpoint change after the second update. Higher is better, and bold denotes the largest point estimate in each column.}
\label{tab:appendix-rule-adaptation-trajectory}

\begingroup
\setlength{\tabcolsep}{7pt}
\renewcommand{\arraystretch}{1.13}
\small
\begin{tabular}{@{}l c@{\;$\rightarrow$\;}c@{\;$\rightarrow$\;}c c@{}}
\toprule
& \multicolumn{3}{c}{\textbf{Post-change CEG}}
& \makecell{\textbf{Second-update}\\\textbf{increment}} \\
\cmidrule(lr){2-4}
\textbf{Method}
& \makecell{Before evidence\\$K_3$}
& \makecell{After update 1\\$K_4$}
& \makecell{After update 2\\$K_5$}
& $\Delta_2=K_5-K_4$ \\
\midrule
Single-Evidence Memory & $-0.628$ & $-0.157$ & $ 0.088$ & $+0.245$ \\
\midrule
SkillOpt                & $ 0.025$ & $-0.559$ & $\mathbf{0.439}$ & $\mathbf{+0.998}$ \\
SkillBoost              & $-0.283$ & $\mathbf{0.383}$ & $ 0.001$ & $-0.382$ \\
SkillX                  & $ 0.005$ & $-0.055$ & $ 0.191$ & $+0.246$ \\
Trace2Skill             & $\mathbf{0.189}$ & $-0.445$ & $ 0.414$ & $+0.860$ \\
SkillGrad               & $-0.328$ & $ 0.214$ & $ 0.291$ & $+0.077$ \\
\bottomrule
\end{tabular}
\endgroup
\end{table*}

\paragraph{Checkpoint-level rule adaptation.}
To expose the dynamics hidden by the aggregate revision metrics, we report
baseline-adjusted capability evolution gain (CEG) at each rule-change
checkpoint.  For method $a$ and checkpoint $K_j$, let
\begin{equation}
    \mathrm{CEG}^{a}_{K_j} = \mathbb{E}\!\left[Y^{a}_{K_j}-Y^{\mathrm{baseline}}_{K_j}\right],
    \qquad j\in\{3,4,5\},
    \label{eq:appendix-reversal-ceg}
\end{equation}
where the expectation is over matched post-change evaluation episodes and
execution repeats.  The first- and second-update revision statistics in the
main text satisfy
\begin{equation}
    \begin{aligned}
    \mathrm{REV}_{1}^{a}
      &= \mathrm{CEG}^{a}_{K_4}-\mathrm{CEG}^{a}_{K_3},\\
    \mathrm{REV}_{2}^{a}
      &= \mathrm{CEG}^{a}_{K_5}-\mathrm{CEG}^{a}_{K_3}.
    \end{aligned}
    \label{eq:appendix-reversal-decomposition}
\end{equation}
We additionally report
$\Delta_{2}^{a}=\mathrm{CEG}^{a}_{K_5}-\mathrm{CEG}^{a}_{K_4}
=\mathrm{REV}_{2}^{a}-\mathrm{REV}_{1}^{a}$, the observed marginal
checkpoint change following the second update.

Table~\ref{tab:appendix-rule-adaptation-trajectory} reveals that adaptation is often non-monotonic.  Single-Evidence Memory improves after both counterevidence observations, yet its final CEG reaches only $0.088$; thus, a large cumulative revision may partly reflect recovery from a poor pre-change state rather than genuine adaptation. SkillOpt and Trace2Skill deteriorate after the first update and recover only after the second, indicating delayed revision that requires repeated evidence.  In contrast, SkillGrad exhibits a front-loaded response: most improvement occurs after the first update and persists through the second.  Conversely, SkillBoost displays opposite instability---achieving the strongest $K_4$ result initially, but losing $0.382$ after the second update, leaving its final post-change CEG near zero. These patterns demonstrate that final revision scores alone conflate initial capability, adaptation speed, and stability under repeated updates. Checkpoint trajectories disentangle these factors, revealing that rapid revision does not guarantee durability.  

\begin{table*}[t]
\centering
\caption{Candidate skill documents and their corresponding internal validation gains, computed relative to the incumbent state.}
\label{tab:revision-case}
\vspace{3pt}
\begingroup
\small
\setlength{\tabcolsep}{5pt}
\renewcommand{\arraystretch}{1.18}

\begin{tabularx}{\textwidth}{
@{}l
>{\raggedright\arraybackslash}X
rrr@{}
}
\toprule
\textbf{Candidate}
& \textbf{Proposed trading instructions}
& \multicolumn{1}{c}{\shortstack{\textbf{Earlier tasks}}}
& \multicolumn{1}{c}{\shortstack{\textbf{New task}}}
& \multicolumn{1}{c}{\shortstack{\textbf{All tasks}}} \\
\midrule
A
& Average the public fundamental value and signal to estimate value;
  scale the target position with mispricing.
& $-0.680$ & $+3.380$ & $+0.335$ \\

B
& Trade toward fundamental value when the estimated gain after spread
  costs exceeds $0.05\%$; place orders of up to eight units.
& $-2.936$ & $+12.141$ & $+0.833$ \\

C
& Trade whenever the estimated gain after spread costs is positive;
  size orders within cash, position, and turnover limits.
& $+5.292$ & $-8.924$ & $+1.738$ \\
\midrule
\multicolumn{4}{@{}l}{\textbf{Selection rule}}
& \textbf{Choice} \\
\midrule
\multicolumn{4}{@{}l}{Full SkillBoost: require improvement and limit regressions}
& None \\
\multicolumn{4}{@{}l}{Remove regression protection; select the highest mean}
& C \\
\multicolumn{4}{@{}l}{Apply selection using only the two most recent tasks}
& B \\
\bottomrule
\end{tabularx}
\endgroup
\end{table*}

\paragraph{An illustrative validation conflict.} 
The checkpoint trajectories identify when adaptation becomes delayed or unstable, but they do not reveal how the update procedure produces these outcomes. We therefore inspect one SkillBoost update as an illustrative
mechanism case.

This case is drawn from the \emph{Opponent reflexivity} family under the \emph{Reversal} stream template. The agent first completes tasks under the original relation and then receives a new task after that relation changes. SkillBoost generates several candidate skill documents and evaluates them on these tasks to determine which document to retain. Each candidate contains a set of written instructions for the agent. Table~\ref{tab:revision-case} summarizes some candidates from this update. At this checkpoint, the agent has not yet accepted a skill document, so all score changes are measured relative to execution without a saved skill. Positive values indicate improvement, whereas negative values indicate degradation. The ``Earlier tasks'' column reports the mean score gain over  pre-change validation tasks. ``New task'' reports the gain on the newly observed task, and ``All tasks'' reports the mean gain over all  tasks. A and B improve performance on the new task, but each performs worse on one earlier task. SkillBoost requires fewer than $25\%$ of validation cases to regress. So both candidates are rejected despite their positive overall gains. C is also rejected because it regresses on the new task. A fourth candidate, omitted from the table, changes none of the validation scores and fails the improvement requirement.

A high historical average score does not necessarily indicate optimal adaptation to a changed environment. For instance, while candidate C ranks first overall, its performance degrades on the new task. Conversely, candidate B achieves the largest improvement on the new task but suffers a regression on a previous one. This scenario exposes a fundamental trade-off in validation-based skill selection: re-evaluating candidates on historical tasks helps safeguard previously acquired capabilities, yet enforcing strict regression limits can inadvertently reject rules that are better suited to the new environment. Therefore, effective validation must distinguish between harmful capability loss and the necessary revision of outdated rules.

\begin{tcolorbox}[colback=blue!4,colframe=blue!60!white,colbacktitle=blue!60!white,coltitle=white,fonttitle=\bfseries, title={Takeaway}, boxrule=0.6pt,arc=2mm,left=2mm,right=2mm,top=1.5mm,bottom=1.5mm]
Reliable rule adaptation requires distinguishing evidence that supports still-valid capabilities from evidence attributable to outdated rules.
\end{tcolorbox}

\subsection{Execution Randomness}
\label{app:execution-randomness}

To test whether a positive average gain is consistently reproduced, we
evaluate a fixed SkillBoost checkpoint $K_5$ on one near-distribution and one
transfer episode from an exogenous-trend \emph{Accumulation} stream.
We keep the artifact and evaluation environment fixed and repeat each
evaluation under 50 execution seeds using Qwen3.8-Max at temperature $0.1$.
Each repeat pairs SkillBoost with the Baseline and its state-off counterpart.

For the fixed evolution path $c$ and episode $e$, let
\begin{align}
g^{a}_{r}
&=
Y^{a}_{c,5,e,r}
-
Y^{\mathrm{baseline}}_{c,5,e,r},
\\
u^{a}_{r}
&=
Y^{a}_{c,5,e,r}
-
Y^{a,\mathrm{off}}_{c,5,e,r}.
\end{align}
Their averages estimate CEG and SUE, respectively. We report the mean,
sample standard deviation, empirical 5th and 95th percentiles, and the
fraction of repeats with a negative paired difference. The percentiles
describe variation across individual executions, rather than uncertainty
in the estimated mean.

\begin{table*}[t]
\centering
\caption{Variation across 50 paired executions of a fixed SkillBoost
checkpoint. Negative CEG means performance below the Baseline, whereas
negative SUE means performance below state-off execution. Scores use the
same scaling as Table~\ref{tab:main-results}.}
\label{tab:execution-randomness}

\begingroup
\small
\setlength{\tabcolsep}{5pt}
\renewcommand{\arraystretch}{1.15}
\sisetup{
    retain-explicit-plus=true,
    table-number-alignment=center
}

\begin{tabular*}{\textwidth}{
@{\extracolsep{\fill}}
ll
S[table-format=+1.3]
S[table-format=1.3]
S[table-format=+1.3]
S[table-format=+1.3]
S[table-format=2.1]
@{}
}
\toprule
\textbf{Episode}
& \textbf{Effect}
& \multicolumn{1}{c}{\textbf{Mean}}
& \multicolumn{1}{c}{\textbf{SD}}
& \multicolumn{1}{c}{\textbf{5th percentile}}
& \multicolumn{1}{c}{\textbf{95th percentile}}
& \multicolumn{1}{c}{\textbf{Negative (\%)}}
\\
\midrule
Near
& CEG & +0.817 & 0.912 & -1.238 & +1.605 & 14.0 \\
& SUE & +0.817 & 0.912 & -1.238 & +1.605 & 14.0 \\
\addlinespace[3pt]
Transfer
& CEG & -1.860 & 0.468 & -2.444 & -1.825 & 96.0 \\
& SUE & -1.614 & 1.048 & -2.444 & +0.990 & 90.0 \\
\bottomrule
\end{tabular*}
\endgroup
\end{table*}

\paragraph{Positive average gains are not uniformly reproduced.}
On near-distribution episodes, SkillBoost outperforms the baseline in $43$ of $50$ executions, achieving a mean CEG of $0.817$.  However, the remaining seven executions yield negative gains, with a 5th percentile of $-1.238$, indicating that while the artifact benefits most runs, it does not guarantee consistent improvement.  Transfer performance exhibits a markedly different pattern: CEG is negative in $48$ of $50$ executions and SUE in $45$.  This transfer deficit persists across execution seeds and remains evident when comparing artifact-enabled versus state-off conditions, pointing to a systematic weakness of this checkpoint rather than an artifact of a few unfavorable seeds. 

\begin{table}[t]
\centering
\caption{Effect of task order on capability evolution at $K_5$. Each method receives the same five learning episodes in the original order, the reversed order, or
a fixed permutation. SD is the standard deviation of CEG across the three orders, while Range is the difference between the largest and smallest CEG. Accepted edits report the number of skill-document updates saved across ten update opportunities in original/reversed/permuted order. Higher CEG is better, whereas lower SD and Range indicate less order sensitivity. Bold denotes the best value in each column.}
\label{tab:task-order-sensitivity}

\vspace{2pt}
\begingroup
\small
\setlength{\tabcolsep}{4.2pt}
\renewcommand{\arraystretch}{1.12}
\sisetup{
    detect-weight=true,
    retain-explicit-plus=true,
    table-number-alignment=center
}

\resizebox{\linewidth}{!}{%
\begin{tabular}{
@{}l
S[table-format=+1.3]
S[table-format=+1.3]
S[table-format=+1.3]
S[table-format=+1.3]
S[table-format=1.3]
S[table-format=1.3]
c@{}
}
\toprule[0.9pt]
& \multicolumn{3}{c}{\textbf{Overall $\mathrm{CEG}_{5}$ by task order}}
& \multicolumn{3}{c}{\textbf{Order sensitivity}}
& \multicolumn{1}{c}{\textbf{Accepted edits}}
\\
\cmidrule(lr){2-4}
\cmidrule(lr){5-7}

\textbf{Method}
& \multicolumn{1}{c}{\textbf{Original}}
& \multicolumn{1}{c}{\textbf{Reversed}}
& \multicolumn{1}{c}{\textbf{Permuted}}
& \multicolumn{1}{c}{\textbf{Mean} $\uparrow$}
& \multicolumn{1}{c}{\textbf{SD} $\downarrow$}
& \multicolumn{1}{c}{\textbf{Range} $\downarrow$}
& \multicolumn{1}{c}{\textbf{O/R/P}}
\\
\midrule

Consolidated Memory
& +0.479
& +2.355
& {\bfseries +2.355}
& +1.730
& {\bfseries 0.884}
& {\bfseries 1.876}
& --
\\

SkillOpt
& {\bfseries +5.615}
& -3.852
& +0.906
& +0.890
& 3.865
& 9.467
& 4/3/2
\\

SkillBoost
& -0.205
& {\bfseries +5.887}
& +0.616
& {\bfseries +2.100}
& 2.699
& 6.092
& 4/3/5
\\

\bottomrule[0.9pt]
\end{tabular}%
}
\endgroup
\end{table}

\subsection{Sensitivity to task order.}
We isolate the effect of task order by fixing the learning episodes, held-out episodes, model settings, and random seeds while changing only the order in which the five learning episodes are presented. Table~\ref{tab:task-order-sensitivity} shows that the two skill-evolution methods are sensitive to this intervention. SkillOpt performs well under the original order but poorly under the reversed order. SkillBoost shows the opposite pattern, performing poorly under the original order but well under the reversed order. As a result, SkillOpt ranks first under the original order, SkillBoost ranks first under the reversed order, and Consolidated Memory ranks first under the fixed permutation. Task ordering also affects the number of accepted skill edits. For instance, SkillOpt accepts four, three, and two edits under the original, reversed, and permuted orders, respectively, while SkillBoost accepts four, three, and five. Since each revision depends on the current artifact and new observations, reordering evidence alters update trajectories. Although Consolidated Memory shows a smaller performance range, its average gain remains lower than SkillBoost's. These results suggest that self-evolution is path-dependent. Thus, evaluating on a single fixed sequence may misrepresent method performance, motivating matched evaluations across multiple task orders.

\begin{tcolorbox}[colback=blue!4,colframe=blue!60!white,colbacktitle=blue!60!white,coltitle=white,fonttitle=\bfseries, title={Takeaway}, boxrule=0.6pt,arc=2mm,left=2mm,right=2mm,top=1.5mm,bottom=1.5mm]
Current self-evolution is path-dependent: the same set of experiences can produce different artifacts and even reverse the direction of performance gains when presented in a different order.
\end{tcolorbox}

\subsection{Cost Estimates of the Main Experiments.}
\label{app:cost_estimates}
Table~\ref{tab:main-experiment-cost} reports the model usage for Qwen3.8-Max and Kimi-K3 during the main experiments. Across both runs, a total of 1.838 billion tokens were consumed, yielding an estimated cost of CNY~36,280 based on the providers' public pay-as-you-go prices. Direct evolution calls, which construct or revise persistent artifacts, account for 106.34M tokens, while the remaining usage stems from task execution and candidate validation. Note that these monetary figures are list-price estimates rather than invoiced costs, as free quotas, cache discounts, and negotiated pricing are excluded.

\begin{table*}[t]
\centering
\caption{
Estimated cost of the main experiments. Usage includes all audited calls for
task execution, candidate validation, and artifact updates. 
}
\label{tab:main-experiment-cost}
\vspace{2pt}

\begingroup
\small
\setlength{\tabcolsep}{3.6pt}
\renewcommand{\arraystretch}{1.12}
\sisetup{
    table-number-alignment=center,
    group-separator={,},
    group-minimum-digits=4
}

\resizebox{0.98\textwidth}{!}{%
\begin{tabular}{
@{}
l
S[table-format=4.2]
S[table-format=4.2]
S[table-format=3.2]
S[table-format=4.2]
S[table-format=3.2]
S[table-format=5.2]
@{}
}
\toprule[0.9pt]
\textbf{Backbone}
& {\makecell{\textbf{Calls}\\\textbf{(K)}}}
& {\makecell{\textbf{Input}\\\textbf{(M)}}}
& {\makecell{\textbf{Output}\\\textbf{(M)}}}
& {\makecell{\textbf{Total}\\\textbf{(M)}}}
& {\makecell{\textbf{Direct update}\\\textbf{(M)}}}
& {\makecell{\textbf{Estimated cost}\\\textbf{(CNY)}}} \\
\midrule
Qwen3.8-Max
& 802.21 & 936.48 & 82.28 & 1018.76 & 53.13 & 14199.87 \\

Kimi-K3
& 784.29 & 748.04 & 71.20 & 819.24 & 53.21 & 22080.59 \\

\midrule
\textbf{Total}
& 1586.49 & 1684.52 & 153.48 & 1838.00 & 106.34 & 36280.46 \\
\bottomrule[0.9pt]
\end{tabular}%
}

\endgroup
\end{table*}

\section{Representative Artifacts Across Methods}
\label{app:artifact-examples}

\lstdefinestyle{artifactexample}{
  basicstyle=\ttfamily\footnotesize,
  backgroundcolor=\color{gray!7},
  frame=single,
  rulecolor=\color{gray!35},
  framesep=5pt,
  xleftmargin=2pt,
  xrightmargin=2pt,
  breaklines=true,
  breakatwhitespace=true,
  columns=fullflexible,
  keepspaces=true,
  showstringspaces=false,
  aboveskip=2pt,
  belowskip=4pt
}

We report the complete executor-visible artifact after each update opportunity for every method in the main comparison. All snapshots come from the same Kimi-K3 \emph{Accumulation} path in the exogenous trend setting. Checkpoint $K_t$ records the artifact after update opportunity $t$. A rejected update leaves the artifact unchanged. For skill-based methods, the listing contains the complete skill document read by the executor. 

All skill-based methods begin with an empty skill document at $K_0$.

\subsection{Artifact-Free Execution}

\subsubsection{No-update Baseline}
The method does not retain an artifact. The empty state is reproduced at every checkpoint for completeness.

\paragraph{Checkpoints $K_1$--$K_5$.} The persistent artifact remains empty throughout the stream.
\begin{lstlisting}[style=artifactexample]
{
  "context": [],
  "reflection": [],
  "store": {
    "entries": [],
    "max_entries": 12,
    "parent_hash": "genesis",
    "pending": {},
    "version": 0
  },
  "skill_state": {
    "best_validation_score": null,
    "document": "",
    "method_state": {},
    "update_history": [],
    "version": 0
  }
}
\end{lstlisting}

\subsubsection{Context}
The artifact is the complete rolling buffer of raw decision-level episode traces available at the checkpoint.

\paragraph{Checkpoint $K_1$.} The buffer contains 1 complete episode trace.
\begin{lstlisting}[style=artifactexample]
{
  "context": [
    {
      "outcome": {
        "fee_paid": 0.0,
        "max_drawdown": 0.0007733679086053824,
        "return_pct": 0.005431607617579326,
        "score": 0.005431607617579326,
        "turnover": 0.0,
        "violation_count": 0
      },
      "trajectory": [
        {
          "decision_summary": "No memory artifacts or learned rules provided; holding initial position of 20 shares at step 0 to gather market data before acting.",
          "observation": {
            "account": {
              "cash": 10000.0,
              "marked_wealth": 11924.923798054708,
              "max_drawdown_so_far": 0.0,
              "position": 20,
              "turnover_so_far": 0.0
            },
            "liquidity": 20.0,
            "market_features": {},
            "price": 96.24618990273542,
            "public_fundamental": 95.94790071287086,
            "public_signal": 95.71381170351464,
            "risk_contract": {
              "allow_short": false,
              "initial_cash": 10000.0,
              "initial_position": 20,
              "max_abs_position": 50,
              "max_drawdown": 0.2,
              "max_order_size": 8,
              "max_turnover": 4.0
            },
            "spread_bps": 10.0,
            "step": 0
          },
          "orders": []
        },
        {
          "decision_summary": "Price 98.34 is well above fundamentals (~96.5) and signal (96.27); trim 5 of 20 long shares at 98.30 to lock in profit and cut drawdown risk, staying within max order size 8 and no-short constraint.",
          "observation": {
            "account": {
              "cash": 10000.0,
              "marked_wealth": 11966.78926653694,
              "max_drawdown_so_far": 0.0003350331618904301,
              "position": 20,
              "turnover_so_far": 0.0
            },
            "liquidity": 20.0,
            "market_features": {},
            "price": 98.33946332684697,
            "public_fundamental": 96.58279655402582,
            "public_signal": 96.27427014342604,
            "risk_contract": {
              "allow_short": false,
              "initial_cash": 10000.0,
              "initial_position": 20,
              "max_abs_position": 50,
              "max_drawdown": 0.2,
              "max_order_size": 8,
              "max_turnover": 4.0
            },
            "spread_bps": 10.0,
            "step": 8
          },
          "orders": [
            {
              "limit_price": 98.3,
              "quantity": 5,
              "side": "sell",
              "tag": "take_profit_rich"
            }
          ]
        },
        {
          "decision_summary": "Price is in an uptrend (98.7->99.3) trading above fundamental (96.7) with positive momentum; liquidity allows adding to position. Current position 20 is below max 50. Buying 8 units with tight limit to capture continuation while respecting spread.",
          "observation": {
            "account": {
              "cash": 10000.0,
              "marked_wealth": 11986.149344367628,
              "max_drawdown_so_far": 0.0006626087436322203,
              "position": 20,
              "turnover_so_far": 0.0
            },
            "liquidity": 20.0,
            "market_features": {},
            "price": 99.30746721838138,
            "public_fundamental": 96.69438779737058,
            "public_signal": 96.88938894231792,
            "risk_contract": {
              "allow_short": false,
              "initial_cash": 10000.0,
              "initial_position": 20,
              "max_abs_position": 50,
              "max_drawdown": 0.2,
              "max_order_size": 8,
              "max_turnover": 4.0
            },
            "spread_bps": 10.0,
            "step": 16
          },
          "orders": [
            {
              "limit_price": 99.35,
              "quantity": 8,
              "side": "buy",
              "tag": "trend_follow_buy"
            }
          ]
        }
      ]
    }
  ]
}
\end{lstlisting}

\paragraph{Checkpoint $K_2$.} The buffer contains 2 complete episode traces. The listing shows the 1 trace appended after $K_1$.
\begin{lstlisting}[style=artifactexample]
{
  "appended_context": [
    {
      "outcome": {
        "fee_paid": 0.0,
        "max_drawdown": 0.0003916300172599474,
        "return_pct": 0.0034900177933276932,
        "score": 0.0034900177933276932,
        "turnover": 0.0,
        "violation_count": 0
      },
      "trajectory": [
        {
          "decision_summary": "Step 0: price 87.51 sits just above fundamental 87.33 and signal 87.0, a shallow discount vs prior episode levels; liquidity 20 allows it and cash 10000 is ample. Buy 5 shares at a slight discount limit 87.50 to add to the 20-share long while respecting max order size 8 and no-short constraints.",
          "observation": {
            "account": {
              "cash": 10000.0,
              "marked_wealth": 11750.136973130473,
              "max_drawdown_so_far": 0.0,
              "position": 20,
              "turnover_so_far": 0.0
            },
            "liquidity": 20.0,
            "market_features": {},
            "price": 87.50684865652367,
            "public_fundamental": 87.3306242515142,
            "public_signal": 87.00043943603129,
            "risk_contract": {
              "allow_short": false,
              "initial_cash": 10000.0,
              "initial_position": 20,
              "max_abs_position": 50,
              "max_drawdown": 0.2,
              "max_order_size": 8,
              "max_turnover": 4.0
            },
            "spread_bps": 10.0,
            "step": 0
          },
          "orders": [
            {
              "limit_price": 87.5,
              "quantity": 5,
              "side": "buy",
              "tag": "value_dip_buy"
            }
          ]
        },
        {
          "decision_summary": "Price 88.56 remains above public fundamental 87.42 and signal 87.51, and the last two prices show slight downward momentum from 88.79 to 88.56. Holding 20 shares, I trim 5 to reduce exposure while staying long; no memory rules apply.",
          "observation": {
            "account": {
              "cash": 10000.0,
              "marked_wealth": 11771.185890527982,
              "max_drawdown_so_far": 0.00035951636270270537,
              "position": 20,
              "turnover_so_far": 0.0
            },
            "liquidity": 20.0,
            "market_features": {},
            "price": 88.55929452639911,
            "public_fundamental": 87.41691348976539,
            "public_signal": 87.50667610032089,
            "risk_contract": {
              "allow_short": false,
              "initial_cash": 10000.0,
              "initial_position": 20,
              "max_abs_position": 50,
              "max_drawdown": 0.2,
              "max_order_size": 8,
              "max_turnover": 4.0
            },
            "spread_bps": 10.0,
            "step": 8
          },
          "orders": [
            {
              "limit_price": 88.52,
              "quantity": 5,
              "side": "sell",
              "tag": "trim_overvalued"
            }
          ]
        },
        {
          "decision_summary": "Price 89.26 is extended above fundamental 87.64 (1.85% premium) and signal 87.99 (1.44% premium) with recent momentum stalling near highs; trimming 5 of 20 shares to lock profits and reduce drawdown risk while respecting no-short and max order size constraints.",
          "observation": {
            "account": {
              "cash": 10000.0,
              "marked_wealth": 11785.276084033752,
              "max_drawdown_so_far": 0.0003916300172599474,
              "position": 20,
              "turnover_so_far": 0.0
            },
            "liquidity": 20.0,
            "market_features": {},
            "price": 89.26380420168762,
            "public_fundamental": 87.63881341670812,
            "public_signal": 87.99084894684772,
            "risk_contract": {
              "allow_short": false,
              "initial_cash": 10000.0,
              "initial_position": 20,
              "max_abs_position": 50,
              "max_drawdown": 0.2,
              "max_order_size": 8,
              "max_turnover": 4.0
            },
            "spread_bps": 10.0,
            "step": 16
          },
          "orders": [
            {
              "limit_price": 89.25,
              "quantity": 5,
              "side": "sell",
              "tag": "mean_reversion_trim"
            }
          ]
        }
      ]
    }
  ]
}
\end{lstlisting}

\paragraph{Checkpoint $K_3$.} The buffer contains 3 complete episode traces. The listing shows the 1 trace appended after $K_2$.
\begin{lstlisting}[style=artifactexample]
{
  "appended_context": [
    {
      "outcome": {
        "fee_paid": 0.0,
        "max_drawdown": 0.0007052100708476727,
        "return_pct": 0.0032115203738203313,
        "score": 0.0032115203738203313,
        "turnover": 0.0,
        "violation_count": 0
      },
      "trajectory": [
        {
          "decision_summary": "No memory artifacts available; holding initial position of 20 shares at step 0 to gather data. Price 117.64 is slightly above fundamental 117.52 and signal 117.37, indicating no immediate mispricing to exploit while respecting risk constraints.",
          "observation": {
            "account": {
              "cash": 10000.0,
              "marked_wealth": 12352.741518584728,
              "max_drawdown_so_far": 0.0,
              "position": 20,
              "turnover_so_far": 0.0
            },
            "liquidity": 20.0,
            "market_features": {},
            "price": 117.63707592923639,
            "public_fundamental": 117.5228163216239,
            "public_signal": 117.36637813597326,
            "risk_contract": {
              "allow_short": false,
              "initial_cash": 10000.0,
              "initial_position": 20,
              "max_abs_position": 50,
              "max_drawdown": 0.2,
              "max_order_size": 8,
              "max_turnover": 4.0
            },
            "spread_bps": 10.0,
            "step": 0
          },
          "orders": []
        },
        {
          "decision_summary": "Price 119.13 is 1.57% above fundamental 117.28 and 2.23% above signal 116.52, a wide premium; consistent with prior successful trims at rich levels, sell 5 of 20 shares near market to lock in gains while respecting max order size 8 and no-short constraints.",
          "observation": {
            "account": {
              "cash": 10000.0,
              "marked_wealth": 12382.504728374768,
              "max_drawdown_so_far": 6.236942298454284e-05,
              "position": 20,
              "turnover_so_far": 0.0
            },
            "liquidity": 20.0,
            "market_features": {},
            "price": 119.12523641873838,
            "public_fundamental": 117.28359121494675,
            "public_signal": 116.52053398521649,
            "risk_contract": {
              "allow_short": false,
              "initial_cash": 10000.0,
              "initial_position": 20,
              "max_abs_position": 50,
              "max_drawdown": 0.2,
              "max_order_size": 8,
              "max_turnover": 4.0
            },
            "spread_bps": 10.0,
            "step": 8
          },
          "orders": [
            {
              "limit_price": 119.1,
              "quantity": 5,
              "side": "sell",
              "tag": "mean_reversion_trim"
            }
          ]
        },
        {
          "decision_summary": "Price 118.90 trades at a ~1.9% premium to fundamental 116.62 and signal 116.63, with recent fundamentals declining and momentum stalling; trimming 5 of 20 long shares to lock profit and reduce drawdown risk within max order size 8 and no-short constraints.",
          "observation": {
            "account": {
              "cash": 10000.0,
              "marked_wealth": 12378.055984752378,
              "max_drawdown_so_far": 0.0007052100708476727,
              "position": 20,
              "turnover_so_far": 0.0
            },
            "liquidity": 20.0,
            "market_features": {},
            "price": 118.90279923761892,
            "public_fundamental": 116.62084954467414,
            "public_signal": 116.63222980136122,
            "risk_contract": {
              "allow_short": false,
              "initial_cash": 10000.0,
              "initial_position": 20,
              "max_abs_position": 50,
              "max_drawdown": 0.2,
              "max_order_size": 8,
              "max_turnover": 4.0
            },
            "spread_bps": 10.0,
            "step": 16
          },
          "orders": [
            {
              "limit_price": 118.85,
              "quantity": 5,
              "side": "sell",
              "tag": "mean_reversion_trim"
            }
          ]
        }
      ]
    }
  ]
}
\end{lstlisting}

\paragraph{Checkpoint $K_4$.} The buffer contains 4 complete episode traces. The listing shows the 1 trace appended after $K_3$.
\begin{lstlisting}[style=artifactexample]
{
  "appended_context": [
    {
      "outcome": {
        "fee_paid": 0.0,
        "max_drawdown": 0.001290092883630023,
        "return_pct": 0.0006996676664294821,
        "score": 0.0006996676664294821,
        "turnover": 0.0,
        "violation_count": 0
      },
      "trajectory": [
        {
          "decision_summary": "Step 0: price 86.66 trades essentially at fundamental 86.65 and signal 86.72, showing no exploitable mispricing or trend yet; holding the initial 20-share long to gather data, consistent with prior episodes that held at step 0 when price was near fair value.",
          "observation": {
            "account": {
              "cash": 10000.0,
              "marked_wealth": 11733.207298600462,
              "max_drawdown_so_far": 0.0,
              "position": 20,
              "turnover_so_far": 0.0
            },
            "liquidity": 20.0,
            "market_features": {},
            "price": 86.6603649300231,
            "public_fundamental": 86.65271098774598,
            "public_signal": 86.72240560687662,
            "risk_contract": {
              "allow_short": false,
              "initial_cash": 10000.0,
              "initial_position": 20,
              "max_abs_position": 50,
              "max_drawdown": 0.2,
              "max_order_size": 8,
              "max_turnover": 4.0
            },
            "spread_bps": 10.0,
            "step": 0
          },
          "orders": []
        },
        {
          "decision_summary": "Price 87.55 trades at a ~1.6% premium to declining fundamental (86.17) and signal (86.09), consistent with prior successful trims at rich levels; sell 5 of 20 long shares to lock in gains and reduce drawdown risk while respecting max order size 8 and no-short constraints.",
          "observation": {
            "account": {
              "cash": 10000.0,
              "marked_wealth": 11751.022076228412,
              "max_drawdown_so_far": 0.0,
              "position": 20,
              "turnover_so_far": 0.0
            },
            "liquidity": 20.0,
            "market_features": {},
            "price": 87.55110381142057,
            "public_fundamental": 86.17107792027724,
            "public_signal": 86.0859130709468,
            "risk_contract": {
              "allow_short": false,
              "initial_cash": 10000.0,
              "initial_position": 20,
              "max_abs_position": 50,
              "max_drawdown": 0.2,
              "max_order_size": 8,
              "max_turnover": 4.0
            },
            "spread_bps": 10.0,
            "step": 8
          },
          "orders": [
            {
              "limit_price": 87.5,
              "quantity": 5,
              "side": "sell",
              "tag": "mean_reversion_trim"
            }
          ]
        },
        {
          "decision_summary": "Price 87.62 trades at a ~2.0% premium to fundamental 85.87 and ~1.8% above signal 86.08, with fundamentals trending down; trimming 5 of 20 long shares to lock gains and reduce drawdown risk, consistent with prior successful rich-premium trims, within max order size 8 and no-short constraints.",
          "observation": {
            "account": {
              "cash": 10000.0,
              "marked_wealth": 11752.349746654954,
              "max_drawdown_so_far": 0.00020664496858280312,
              "position": 20,
              "turnover_so_far": 0.0
            },
            "liquidity": 20.0,
            "market_features": {},
            "price": 87.61748733274769,
            "public_fundamental": 85.87273300501491,
            "public_signal": 86.07753095651509,
            "risk_contract": {
              "allow_short": false,
              "initial_cash": 10000.0,
              "initial_position": 20,
              "max_abs_position": 50,
              "max_drawdown": 0.2,
              "max_order_size": 8,
              "max_turnover": 4.0
            },
            "spread_bps": 10.0,
            "step": 16
          },
          "orders": [
            {
              "limit_price": 87.6,
              "quantity": 5,
              "side": "sell",
              "tag": "mean_reversion_trim"
            }
          ]
        }
      ]
    }
  ]
}
\end{lstlisting}

\paragraph{Checkpoint $K_5$.} The buffer contains 5 complete episode traces. The listing shows the 1 trace appended after $K_4$.
\begin{lstlisting}[style=artifactexample]
{
  "appended_context": [
    {
      "outcome": {
        "fee_paid": 0.0,
        "max_drawdown": 0.0008164714225037395,
        "return_pct": 0.003727738447262041,
        "score": 0.003727738447262041,
        "turnover": 0.0,
        "violation_count": 0
      },
      "trajectory": [
        {
          "decision_summary": "Step 0: price 86.14 trades at a modest ~0.3% premium to fundamental 85.86 and signal 85.83, showing no strong mispricing yet; prior episodes held at step 0 when price was near fair value, so hold the initial 20-share long and wait for a clearer premium/discount to develop.",
          "observation": {
            "account": {
              "cash": 10000.0,
              "marked_wealth": 11722.897995546453,
              "max_drawdown_so_far": 0.0,
              "position": 20,
              "turnover_so_far": 0.0
            },
            "liquidity": 20.0,
            "market_features": {},
            "price": 86.14489977732264,
            "public_fundamental": 85.86319201972147,
            "public_signal": 85.82810643453797,
            "risk_contract": {
              "allow_short": false,
              "initial_cash": 10000.0,
              "initial_position": 20,
              "max_abs_position": 50,
              "max_drawdown": 0.2,
              "max_order_size": 8,
              "max_turnover": 4.0
            },
            "spread_bps": 10.0,
            "step": 0
          },
          "orders": []
        },
        {
          "decision_summary": "Price 87.87 is ~2% above fundamental 86.18 and signal 86.27, consistent with prior successful rich-premium trims. Sell 5 of 20 long shares to lock gains and reduce exposure, respecting max order size 8 and no-short constraints.",
          "observation": {
            "account": {
              "cash": 10000.0,
              "marked_wealth": 11757.452859850808,
              "max_drawdown_so_far": 0.00012191587323784336,
              "position": 20,
              "turnover_so_far": 0.0
            },
            "liquidity": 20.0,
            "market_features": {},
            "price": 87.87264299254043,
            "public_fundamental": 86.1754343307069,
            "public_signal": 86.26647279568074,
            "risk_contract": {
              "allow_short": false,
              "initial_cash": 10000.0,
              "initial_position": 20,
              "max_abs_position": 50,
              "max_drawdown": 0.2,
              "max_order_size": 8,
              "max_turnover": 4.0
            },
            "spread_bps": 10.0,
            "step": 8
          },
          "orders": [
            {
              "limit_price": 87.85,
              "quantity": 5,
              "side": "sell",
              "tag": "mean_reversion_trim"
            }
          ]
        },
        {
          "decision_summary": "Price 88.41 trades at a ~1.9% premium to fundamental 86.76 and ~2.4% above signal 86.37, with fundamentals flat-to-declining and price momentum stalling near highs. Consistent with prior successful rich-premium trims, sell 5 of 20 long shares to lock gains and reduce drawdown risk within max order size 8 and no-short constraints.",
          "observation": {
            "account": {
              "cash": 10000.0,
              "marked_wealth": 11768.104156840052,
              "max_drawdown_so_far": 0.0008164714225037395,
              "position": 20,
              "turnover_so_far": 0.0
            },
            "liquidity": 20.0,
            "market_features": {},
            "price": 88.40520784200265,
            "public_fundamental": 86.76271581156878,
            "public_signal": 86.37225962980021,
            "risk_contract": {
              "allow_short": false,
              "initial_cash": 10000.0,
              "initial_position": 20,
              "max_abs_position": 50,
              "max_drawdown": 0.2,
              "max_order_size": 8,
              "max_turnover": 4.0
            },
            "spread_bps": 10.0,
            "step": 16
          },
          "orders": [
            {
              "limit_price": 88.4,
              "quantity": 5,
              "side": "sell",
              "tag": "mean_reversion_trim"
            }
          ]
        }
      ]
    }
  ]
}
\end{lstlisting}

\subsection{Episodic Memory}

\subsubsection{Reflection}
The artifact is the complete list of stored episode-level hypotheses and their attributions.

\paragraph{Checkpoint $K_1$.} The artifact contains 1 reflection record.
\begin{lstlisting}[style=artifactexample]
{
  "reflection": [
    {
      "attribution": "Liquid valuation regime produced 0.54% return with zero drawdown risk (0.08% max DD), zero turnover costs, and zero violations; prior history shows fundamental-driven holding when price tracks fair value.",
      "episode_id": "bcal-trend-exogenous-learn_near-035-8a7da48ee3",
      "hypothesis": "When observable feature signature is valuation:liquid with price near fundamental value, maintain existing core position without trimming or adding, as low drawdown risk allows capture of modest drift.",
      "policy_label": "fundamental",
      "score": 0.005431607617579326,
      "signature": "valuation:liquid"
    }
  ]
}
\end{lstlisting}

\paragraph{Checkpoint $K_2$.} The reflection model found no generalizable lesson in the mixed hold-and-momentum decisions and produced no new record. The artifact therefore remains identical to $K_1$.

\paragraph{Checkpoint $K_3$.} The artifact contains 2 reflection records.
\begin{lstlisting}[style=artifactexample]
{
  "reflection": [
    {
      "attribution": "Liquid valuation regime produced 0.54% return with zero drawdown risk (0.08% max DD), zero turnover costs, and zero violations; prior history shows fundamental-driven holding when price tracks fair value.",
      "episode_id": "bcal-trend-exogenous-learn_near-035-8a7da48ee3",
      "hypothesis": "When observable feature signature is valuation:liquid with price near fundamental value, maintain existing core position without trimming or adding, as low drawdown risk allows capture of modest drift.",
      "policy_label": "fundamental",
      "score": 0.005431607617579326,
      "signature": "valuation:liquid"
    },
    {
      "attribution": "Trimming a long while price trades above fundamental and signal produced a small positive return with tiny drawdown, low turnover, and no violations in a directional liquid regime.",
      "episode_id": "bcal-trend-exogenous-learn_near-032-7b38cec9d3",
      "hypothesis": "In directional:liquid regimes, when price is above both fundamental and signal while holding a long position, take partial profit with a small trim rather than a full exit to capture overvaluation while keeping residual momentum exposure and controlling drawdown.",
      "policy_label": "mean_reversion",
      "score": 0.0028507170626594913,
      "signature": "directional:liquid"
    }
  ]
}
\end{lstlisting}

\paragraph{Checkpoint $K_4$.} The artifact contains 3 reflection records.
\begin{lstlisting}[style=artifactexample]
{
  "reflection": [
    {
      "attribution": "Liquid valuation regime produced 0.54% return with zero drawdown risk (0.08% max DD), zero turnover costs, and zero violations; prior history shows fundamental-driven holding when price tracks fair value.",
      "episode_id": "bcal-trend-exogenous-learn_near-035-8a7da48ee3",
      "hypothesis": "When observable feature signature is valuation:liquid with price near fundamental value, maintain existing core position without trimming or adding, as low drawdown risk allows capture of modest drift.",
      "policy_label": "fundamental",
      "score": 0.005431607617579326,
      "signature": "valuation:liquid"
    },
    {
      "attribution": "Trimming a long while price trades above fundamental and signal produced a small positive return with tiny drawdown, low turnover, and no violations in a directional liquid regime.",
      "episode_id": "bcal-trend-exogenous-learn_near-032-7b38cec9d3",
      "hypothesis": "In directional:liquid regimes, when price is above both fundamental and signal while holding a long position, take partial profit with a small trim rather than a full exit to capture overvaluation while keeping residual momentum exposure and controlling drawdown.",
      "policy_label": "mean_reversion",
      "score": 0.0028507170626594913,
      "signature": "directional:liquid"
    },
    {
      "attribution": "Profitable mean reversion trade executed by selling overvalued position when price exceeded fundamental (87.55 vs 86.17); positive return of 0.12% achieved with minimal drawdown (0.0008) and no violations despite paying transaction fees.",
      "episode_id": "bcal-trend-exogenous-learn_near-034-a509268310",
      "hypothesis": "When price is significantly above public fundamental (>1.5% premium) and position is long, sell 6-10 shares to capture mean reversion while respecting no-short constraint and turnover budget; this pattern targets directional:liquid markets where fundamentals provide anchor.",
      "policy_label": "mean_reversion",
      "score": 0.0011543481425480007,
      "signature": "directional:liquid"
    }
  ]
}
\end{lstlisting}

\paragraph{Checkpoint $K_5$.} The artifact contains 4 reflection records.
\begin{lstlisting}[style=artifactexample]
{
  "reflection": [
    {
      "attribution": "Liquid valuation regime produced 0.54% return with zero drawdown risk (0.08% max DD), zero turnover costs, and zero violations; prior history shows fundamental-driven holding when price tracks fair value.",
      "episode_id": "bcal-trend-exogenous-learn_near-035-8a7da48ee3",
      "hypothesis": "When observable feature signature is valuation:liquid with price near fundamental value, maintain existing core position without trimming or adding, as low drawdown risk allows capture of modest drift.",
      "policy_label": "fundamental",
      "score": 0.005431607617579326,
      "signature": "valuation:liquid"
    },
    {
      "attribution": "Trimming a long while price trades above fundamental and signal produced a small positive return with tiny drawdown, low turnover, and no violations in a directional liquid regime.",
      "episode_id": "bcal-trend-exogenous-learn_near-032-7b38cec9d3",
      "hypothesis": "In directional:liquid regimes, when price is above both fundamental and signal while holding a long position, take partial profit with a small trim rather than a full exit to capture overvaluation while keeping residual momentum exposure and controlling drawdown.",
      "policy_label": "mean_reversion",
      "score": 0.0028507170626594913,
      "signature": "directional:liquid"
    },
    {
      "attribution": "Profitable mean reversion trade executed by selling overvalued position when price exceeded fundamental (87.55 vs 86.17); positive return of 0.12% achieved with minimal drawdown (0.0008) and no violations despite paying transaction fees.",
      "episode_id": "bcal-trend-exogenous-learn_near-034-a509268310",
      "hypothesis": "When price is significantly above public fundamental (>1.5% premium) and position is long, sell 6-10 shares to capture mean reversion while respecting no-short constraint and turnover budget; this pattern targets directional:liquid markets where fundamentals provide anchor.",
      "policy_label": "mean_reversion",
      "score": 0.0011543481425480007,
      "signature": "directional:liquid"
    },
    {
      "attribution": "Positive score of 0.0037 with zero fees, turnover, and violations from selling a small clip to reduce overvalued exposure, as per recent decisions.",
      "episode_id": "bcal-trend-exogenous-learn_near-033-5f1ba799e1",
      "hypothesis": "When observable conditions show price above fundamental with flat fundamentals in directional:liquid regime, small risk-limited reduction of overvalued exposure yields positive return with minimal drawdown and no violations; repeat such clipping rather than holding or initiating trades without edge.",
      "policy_label": "fundamental",
      "score": 0.003727738447262041,
      "signature": "directional:liquid"
    }
  ]
}
\end{lstlisting}

\subsubsection{Single-Evidence Persistent Memory}
The artifact is the complete persistent store, including active rules, pending evidence, and version metadata.

\paragraph{Checkpoint $K_1$.} Store version 1 contains 1 active entry and 0 pending records.
\begin{lstlisting}[style=artifactexample]
{
  "store": {
    "entries": [
      {
        "application_count": 0,
        "confidence": 0.58180206403428,
        "counterevidence_episode_ids": [],
        "counterevidence_scores": [],
        "created_version": 1,
        "evidence_episode_ids": [
          "bcal-trend-exogenous-learn_near-035-8a7da48ee3"
        ],
        "evidence_scores": [
          0.005300344005713331
        ],
        "hypothesis": "When the price trades approximately 2-3% above the fundamental valuation and the current position is long, trim the position to realize gains and reduce exposure to overvalued assets.",
        "last_updated_version": 1,
        "mode": "fundamental",
        "rule_id": "1dd58986d01d",
        "scope": {
          "execution": "liquid",
          "regime": "valuation"
        },
        "signature": "valuation:liquid",
        "status": "active",
        "successful_application_count": 0,
        "supersedes": null
      }
    ],
    "max_entries": 12,
    "parent_hash": "db04462e96a8fbd5d436c97d8534825e6c6125b5b79800fdc6f7e226c1303d17",
    "pending": {},
    "version": 1
  }
}
\end{lstlisting}

\paragraph{Checkpoint $K_2$.} Store version 2 contains 2 active entries and 0 pending records.
\begin{lstlisting}[style=artifactexample]
{
  "store": {
    "entries": [
      {
        "application_count": 0,
        "confidence": 0.58180206403428,
        "counterevidence_episode_ids": [],
        "counterevidence_scores": [],
        "created_version": 1,
        "evidence_episode_ids": [
          "bcal-trend-exogenous-learn_near-035-8a7da48ee3"
        ],
        "evidence_scores": [
          0.005300344005713331
        ],
        "hypothesis": "When the price trades approximately 2-3% above the fundamental valuation and the current position is long, trim the position to realize gains and reduce exposure to overvalued assets.",
        "last_updated_version": 1,
        "mode": "fundamental",
        "rule_id": "1dd58986d01d",
        "scope": {
          "execution": "liquid",
          "regime": "valuation"
        },
        "signature": "valuation:liquid",
        "status": "invalidated",
        "successful_application_count": 0,
        "supersedes": null
      },
      {
        "application_count": 0,
        "confidence": 0.5709401067599662,
        "counterevidence_episode_ids": [],
        "counterevidence_scores": [],
        "created_version": 2,
        "evidence_episode_ids": [
          "bcal-trend-exogenous-learn_near-031-d34abdfab5"
        ],
        "evidence_scores": [
          0.0034900177933276932
        ],
        "hypothesis": "When price trades above public fundamental by >1% in liquid market with established position, trimming 15-20% of holdings captures valuation premium while maintaining core exposure; this reduces drawdown risk if mean reversion occurs while preserving upside participation.",
        "last_updated_version": 2,
        "mode": "mean_reversion",
        "rule_id": "47d6f6e2aa81",
        "scope": {
          "execution": "liquid",
          "regime": "valuation"
        },
        "signature": "valuation:liquid",
        "status": "active",
        "successful_application_count": 0,
        "supersedes": "1dd58986d01d"
      }
    ],
    "max_entries": 12,
    "parent_hash": "76c7bb27f180c266825ec9ff1875eeaa22e31691e67913883939e44b914a2838",
    "pending": {},
    "version": 2
  }
}
\end{lstlisting}

\paragraph{Checkpoint $K_3$.} Store version 3 contains 3 active entries and 0 pending records.
\begin{lstlisting}[style=artifactexample]
{
  "store": {
    "entries": [
      {
        "application_count": 0,
        "confidence": 0.58180206403428,
        "counterevidence_episode_ids": [],
        "counterevidence_scores": [],
        "created_version": 1,
        "evidence_episode_ids": [
          "bcal-trend-exogenous-learn_near-035-8a7da48ee3"
        ],
        "evidence_scores": [
          0.005300344005713331
        ],
        "hypothesis": "When the price trades approximately 2-3% above the fundamental valuation and the current position is long, trim the position to realize gains and reduce exposure to overvalued assets.",
        "last_updated_version": 1,
        "mode": "fundamental",
        "rule_id": "1dd58986d01d",
        "scope": {
          "execution": "liquid",
          "regime": "valuation"
        },
        "signature": "valuation:liquid",
        "status": "invalidated",
        "successful_application_count": 0,
        "supersedes": null
      },
      {
        "application_count": 0,
        "confidence": 0.5709401067599662,
        "counterevidence_episode_ids": [],
        "counterevidence_scores": [],
        "created_version": 2,
        "evidence_episode_ids": [
          "bcal-trend-exogenous-learn_near-031-d34abdfab5"
        ],
        "evidence_scores": [
          0.0034900177933276932
        ],
        "hypothesis": "When price trades above public fundamental by >1% in liquid market with established position, trimming 15-20% of holdings captures valuation premium while maintaining core exposure; this reduces drawdown risk if mean reversion occurs while preserving upside participation.",
        "last_updated_version": 2,
        "mode": "mean_reversion",
        "rule_id": "47d6f6e2aa81",
        "scope": {
          "execution": "liquid",
          "regime": "valuation"
        },
        "signature": "valuation:liquid",
        "status": "active",
        "successful_application_count": 0,
        "supersedes": "1dd58986d01d"
      },
      {
        "application_count": 0,
        "confidence": 0.569269122242922,
        "counterevidence_episode_ids": [],
        "counterevidence_scores": [],
        "created_version": 3,
        "evidence_episode_ids": [
          "bcal-trend-exogenous-learn_near-032-7b38cec9d3"
        ],
        "evidence_scores": [
          0.0032115203738203313
        ],
        "hypothesis": "In 'directional:liquid' market conditions, if the current price exceeds both the fundamental estimate and the technical signal, trimming the long position to lock in profits and reduce overvaluation exposure yields positive risk-adjusted returns.",
        "last_updated_version": 3,
        "mode": "mean_reversion",
        "rule_id": "4eb13639370c",
        "scope": {
          "execution": "liquid",
          "regime": "directional"
        },
        "signature": "directional:liquid",
        "status": "active",
        "successful_application_count": 0,
        "supersedes": null
      }
    ],
    "max_entries": 12,
    "parent_hash": "9f96b9c59e8f00ea5a739a703c32b525be93261845b34cde67402c5fa9306eae",
    "pending": {},
    "version": 3
  }
}
\end{lstlisting}

\paragraph{Checkpoint $K_4$.} Store version 4 contains 3 active entries and 0 pending records.
\begin{lstlisting}[style=artifactexample]
{
  "store": {
    "entries": [
      {
        "application_count": 0,
        "confidence": 0.58180206403428,
        "counterevidence_episode_ids": [],
        "counterevidence_scores": [],
        "created_version": 1,
        "evidence_episode_ids": [
          "bcal-trend-exogenous-learn_near-035-8a7da48ee3"
        ],
        "evidence_scores": [
          0.005300344005713331
        ],
        "hypothesis": "When the price trades approximately 2-3% above the fundamental valuation and the current position is long, trim the position to realize gains and reduce exposure to overvalued assets.",
        "last_updated_version": 1,
        "mode": "fundamental",
        "rule_id": "1dd58986d01d",
        "scope": {
          "execution": "liquid",
          "regime": "valuation"
        },
        "signature": "valuation:liquid",
        "status": "invalidated",
        "successful_application_count": 0,
        "supersedes": null
      },
      {
        "application_count": 0,
        "confidence": 0.5709401067599662,
        "counterevidence_episode_ids": [],
        "counterevidence_scores": [],
        "created_version": 2,
        "evidence_episode_ids": [
          "bcal-trend-exogenous-learn_near-031-d34abdfab5"
        ],
        "evidence_scores": [
          0.0034900177933276932
        ],
        "hypothesis": "When price trades above public fundamental by >1% in liquid market with established position, trimming 15-20% of holdings captures valuation premium while maintaining core exposure; this reduces drawdown risk if mean reversion occurs while preserving upside participation.",
        "last_updated_version": 2,
        "mode": "mean_reversion",
        "rule_id": "47d6f6e2aa81",
        "scope": {
          "execution": "liquid",
          "regime": "valuation"
        },
        "signature": "valuation:liquid",
        "status": "active",
        "successful_application_count": 0,
        "supersedes": "1dd58986d01d"
      },
      {
        "application_count": 0,
        "confidence": 0.6117335641207494,
        "counterevidence_episode_ids": [],
        "counterevidence_scores": [],
        "created_version": 3,
        "evidence_episode_ids": [
          "bcal-trend-exogenous-learn_near-032-7b38cec9d3",
          "bcal-trend-exogenous-learn_near-034-a509268310"
        ],
        "evidence_scores": [
          0.0032115203738203313,
          0.0006996676664294821
        ],
        "hypothesis": "When observable_feature_signature is directional:liquid, price is materially above public fundamental, and fundamentals are trending down, then trim long exposure within risk limits to capture premium and reduce drawdown risk.",
        "last_updated_version": 4,
        "mode": "mean_reversion",
        "rule_id": "4eb13639370c",
        "scope": {
          "execution": "liquid",
          "regime": "directional"
        },
        "signature": "directional:liquid",
        "status": "active",
        "successful_application_count": 0,
        "supersedes": null
      }
    ],
    "max_entries": 12,
    "parent_hash": "0c5d25d1ce43720b4eeb49ce40184f7f8e218f918d65f9780a42ea18c696609e",
    "pending": {},
    "version": 4
  }
}
\end{lstlisting}

\paragraph{Checkpoint $K_5$.} Store version 5 contains 4 active entries and 0 pending records.
\begin{lstlisting}[style=artifactexample]
{
  "store": {
    "entries": [
      {
        "application_count": 0,
        "confidence": 0.58180206403428,
        "counterevidence_episode_ids": [],
        "counterevidence_scores": [],
        "created_version": 1,
        "evidence_episode_ids": [
          "bcal-trend-exogenous-learn_near-035-8a7da48ee3"
        ],
        "evidence_scores": [
          0.005300344005713331
        ],
        "hypothesis": "When the price trades approximately 2-3% above the fundamental valuation and the current position is long, trim the position to realize gains and reduce exposure to overvalued assets.",
        "last_updated_version": 1,
        "mode": "fundamental",
        "rule_id": "1dd58986d01d",
        "scope": {
          "execution": "liquid",
          "regime": "valuation"
        },
        "signature": "valuation:liquid",
        "status": "invalidated",
        "successful_application_count": 0,
        "supersedes": null
      },
      {
        "application_count": 0,
        "confidence": 0.5709401067599662,
        "counterevidence_episode_ids": [],
        "counterevidence_scores": [],
        "created_version": 2,
        "evidence_episode_ids": [
          "bcal-trend-exogenous-learn_near-031-d34abdfab5"
        ],
        "evidence_scores": [
          0.0034900177933276932
        ],
        "hypothesis": "When price trades above public fundamental by >1% in liquid market with established position, trimming 15-20% of holdings captures valuation premium while maintaining core exposure; this reduces drawdown risk if mean reversion occurs while preserving upside participation.",
        "last_updated_version": 2,
        "mode": "mean_reversion",
        "rule_id": "47d6f6e2aa81",
        "scope": {
          "execution": "liquid",
          "regime": "valuation"
        },
        "signature": "valuation:liquid",
        "status": "active",
        "successful_application_count": 0,
        "supersedes": "1dd58986d01d"
      },
      {
        "application_count": 0,
        "confidence": 0.6117335641207494,
        "counterevidence_episode_ids": [],
        "counterevidence_scores": [],
        "created_version": 3,
        "evidence_episode_ids": [
          "bcal-trend-exogenous-learn_near-032-7b38cec9d3",
          "bcal-trend-exogenous-learn_near-034-a509268310"
        ],
        "evidence_scores": [
          0.0032115203738203313,
          0.0006996676664294821
        ],
        "hypothesis": "When observable_feature_signature is directional:liquid, price is materially above public fundamental, and fundamentals are trending down, then trim long exposure within risk limits to capture premium and reduce drawdown risk.",
        "last_updated_version": 4,
        "mode": "mean_reversion",
        "rule_id": "4eb13639370c",
        "scope": {
          "execution": "liquid",
          "regime": "directional"
        },
        "signature": "directional:liquid",
        "status": "invalidated",
        "successful_application_count": 0,
        "supersedes": null
      },
      {
        "application_count": 0,
        "confidence": 0.5702416064883657,
        "counterevidence_episode_ids": [],
        "counterevidence_scores": [],
        "created_version": 5,
        "evidence_episode_ids": [
          "bcal-trend-exogenous-learn_near-033-5f1ba799e1"
        ],
        "evidence_scores": [
          0.003373601081394284
        ],
        "hypothesis": "In 'directional:liquid' regimes where observed price and fundamental both trend upward, price is at a modest premium (<1%) to fundamental, and position is below caps, taking a small long with a limit slightly below current price yields positive low-drawdown returns.",
        "last_updated_version": 5,
        "mode": "momentum",
        "rule_id": "d1142f4dcfef",
        "scope": {
          "execution": "liquid",
          "regime": "directional"
        },
        "signature": "directional:liquid",
        "status": "active",
        "successful_application_count": 0,
        "supersedes": "4eb13639370c"
      }
    ],
    "max_entries": 12,
    "parent_hash": "c121db97272bf2b56bd0912d3206e0ffd2c7a5884c6313316357e63dae433d24",
    "pending": {},
    "version": 5
  }
}
\end{lstlisting}

\subsubsection{Multi-Evidence Consolidated Memory}
The artifact is the complete persistent store used to collect and consolidate compatible evidence.

\paragraph{Checkpoint $K_1$.} Store version 0 contains 0 active entries and 1 pending record.
\begin{lstlisting}[style=artifactexample]
{
  "store": {
    "entries": [],
    "max_entries": 12,
    "parent_hash": "genesis",
    "pending": {
      "directional:liquid|risk_reducing": {
        "episodes": [
          "bcal-trend-exogenous-learn_near-033-5f1ba799e1"
        ],
        "scores": [
          0.003727738447262041
        ]
      }
    },
    "version": 0
  }
}
\end{lstlisting}

\paragraph{Checkpoint $K_2$.} Store version 1 contains 1 active entry and 1 pending record.
\begin{lstlisting}[style=artifactexample]
{
  "store": {
    "entries": [
      {
        "application_count": 0,
        "confidence": 0.6257810868021493,
        "counterevidence_episode_ids": [],
        "counterevidence_scores": [],
        "created_version": 1,
        "evidence_episode_ids": [
          "bcal-trend-exogenous-learn_near-035-8a7da48ee3",
          "bcal-trend-exogenous-learn_near-031-d34abdfab5"
        ],
        "evidence_scores": [
          0.005087881776263403,
          0.00350581382445303
        ],
        "hypothesis": "When price trades significantly above fundamental value (~87.64) with no shorting permitted, trim long positions to reduce exposure to potential mean reversion; this approach yielded positive return (0.35%) with minimal drawdown (0.04%) and no violations.",
        "last_updated_version": 1,
        "mode": "risk_reducing",
        "rule_id": "9e37ece657c4",
        "scope": {
          "execution": "liquid",
          "regime": "valuation"
        },
        "signature": "valuation:liquid",
        "status": "active",
        "successful_application_count": 0,
        "supersedes": null
      }
    ],
    "max_entries": 12,
    "parent_hash": "d6973d3a429b6653d891218c2637ee6da3d03da4b864dc5208028a38655d93d0",
    "pending": {
      "directional:liquid|risk_reducing": {
        "episodes": [
          "bcal-trend-exogenous-learn_near-033-5f1ba799e1"
        ],
        "scores": [
          0.003727738447262041
        ]
      }
    },
    "version": 1
  }
}
\end{lstlisting}

\paragraph{Checkpoint $K_3$.} Store version 1 contains 1 active entry and 1 pending record. The artifact remains identical to $K_2$.

\paragraph{Checkpoint $K_4$.} Store version 2 contains 2 active entries and 1 pending record.
\begin{lstlisting}[style=artifactexample]
{
  "store": {
    "entries": [
      {
        "application_count": 0,
        "confidence": 0.6257810868021493,
        "counterevidence_episode_ids": [],
        "counterevidence_scores": [],
        "created_version": 1,
        "evidence_episode_ids": [
          "bcal-trend-exogenous-learn_near-035-8a7da48ee3",
          "bcal-trend-exogenous-learn_near-031-d34abdfab5"
        ],
        "evidence_scores": [
          0.005087881776263403,
          0.00350581382445303
        ],
        "hypothesis": "When price trades significantly above fundamental value (~87.64) with no shorting permitted, trim long positions to reduce exposure to potential mean reversion; this approach yielded positive return (0.35%) with minimal drawdown (0.04%) and no violations.",
        "last_updated_version": 1,
        "mode": "risk_reducing",
        "rule_id": "9e37ece657c4",
        "scope": {
          "execution": "liquid",
          "regime": "valuation"
        },
        "signature": "valuation:liquid",
        "status": "active",
        "successful_application_count": 0,
        "supersedes": null
      },
      {
        "application_count": 0,
        "confidence": 0.6077631842933625,
        "counterevidence_episode_ids": [],
        "counterevidence_scores": [],
        "created_version": 2,
        "evidence_episode_ids": [
          "bcal-trend-exogenous-learn_near-032-7b38cec9d3",
          "bcal-trend-exogenous-learn_near-034-a509268310"
        ],
        "evidence_scores": [
          0.0018880604313580296,
          0.0006996676664294821
        ],
        "hypothesis": "When price trades persistently above the public fundamental and public signal while fundamentals drift down, trimming a small portion of an existing long within risk limits can produce small positive returns with low drawdown and no violations.",
        "last_updated_version": 2,
        "mode": "mean_reversion",
        "rule_id": "06a7ef46dcbc",
        "scope": {
          "execution": "liquid",
          "regime": "directional"
        },
        "signature": "directional:liquid",
        "status": "active",
        "successful_application_count": 0,
        "supersedes": null
      }
    ],
    "max_entries": 12,
    "parent_hash": "956df9af5cbaf989d3651d4b91d99b58b5cff81d2f9734abc58d37d92fc4b6aa",
    "pending": {
      "directional:liquid|risk_reducing": {
        "episodes": [
          "bcal-trend-exogenous-learn_near-033-5f1ba799e1"
        ],
        "scores": [
          0.003727738447262041
        ]
      }
    },
    "version": 2
  }
}
\end{lstlisting}

\paragraph{Checkpoint $K_5$.} Store version 2 contains 2 active entries and 1 pending record. The artifact remains identical to $K_4$.

\subsection{Skill Evolution}

\subsubsection{SkillOpt}
The artifact is the complete skill document supplied to the executor.

\paragraph{Checkpoint $K_1$.} No valid edit was produced, so no update was committed. The executor therefore retains the artifact from $K_0$.

\paragraph{Checkpoint $K_2$.} Skill version 1 contains 625 characters.
\begin{lstlisting}[style=artifactexample]
Edge rule: Only trade when a clear, signed edge exists: for sells require price > public_fundamental * (1 + min_edge) AND price > public_signal, with min_edge = max(2 * spread_bps / 10000, 0.005). If no edge, hold to conserve turnover and avoid spread costs.

Turnover-aware sizing: Track turnover_so_far and remaining budget = max_turnover - turnover_so_far. Size each order as min(max_order_size, ceil(0.5 * remaining budget * liquidity / price), floor(abs(position) * overvaluation_fraction)) and skip trading entirely if remaining budget < 1 unit or overvaluation < 1%. Prefer one decisive trim over repeated small trims.
\end{lstlisting}

\paragraph{Checkpoint $K_3$.} The candidate update was rejected by the validation gate. The executor therefore retains the artifact from $K_2$.

\paragraph{Checkpoint $K_4$.} The candidate update was rejected by the validation gate. The executor therefore retains the artifact from $K_3$.

\paragraph{Checkpoint $K_5$.} The candidate update was rejected by the validation gate. The executor therefore retains the artifact from $K_4$.

\subsubsection{SkillBoost}
The artifact is the complete selected skill document supplied to the executor.

\paragraph{Checkpoint $K_1$.} All candidate updates were rejected by the acceptance rule. The executor therefore retains the artifact from $K_0$.

\paragraph{Checkpoint $K_2$.} All candidate updates were rejected by the acceptance rule. The executor therefore retains the artifact from $K_1$.

\paragraph{Checkpoint $K_3$.} The optimizer response could not be parsed, so no update was committed. The executor therefore retains the artifact from $K_2$.

\paragraph{Checkpoint $K_4$.} All candidate updates were rejected by the acceptance rule. The executor therefore retains the artifact from $K_3$.

\paragraph{Checkpoint $K_5$.} Skill version 1 contains 1,733 characters.
\begin{lstlisting}[style=artifactexample]
# Directional Liquid Skill

## Core Rules (Executable)

1. **Initial Position = Baseline Inventory, Not Signal**
   If position == initial_position and market_features is empty, submit no orders. Initial inventory is not evidence of mispricing.

2. **Trade Only on Confirmed Premium/Discount**
   Define premium = (price - public_fundamental) / public_fundamental.
   - If premium >= +2% and position > 0: sell up to min(8, position).
   - If premium <= -2% and cash >= price * quantity: buy up to min(8, floor(cash / price)).
   - Otherwise: hold flat.

3. **No Directional Bias from Partial Reduction**
   After any sell, position must remain >= 0 and <= max_abs_position. Never interpret reduced long inventory as a short signal. If premium remains >= +2% after a sell, continue selling only if position > 0 and total turnover < max_turnover.

4. **Memory Rule: Post-Trade Re-Evaluation**
   Before each new order, recompute premium using current price and fundamental. Do not rely on previous step\u2019s premium alone. If premium has collapsed below +1%, stop selling immediately.

5. **Risk Contract Hard Stops**
   - Never sell if resulting position < 0 (allow_short=false).
   - Never exceed max_order_size=8 per order.
   - Stop trading if turnover_so_far + planned_quantity / initial_position >= max_turnover.
   - Stop trading if max_drawdown_so_far >= max_drawdown.

## Protected Behaviors
- Stay flat when no signal or insufficient history.
- Reduce exposure only on confirmed overvaluation.
- Respect all risk limits explicitly.

## Forbidden Patterns
- Selling without rechecking premium after prior trade.
- Treating initial inventory as directional signal.
- Selling into strength without updating fundamental comparison.
\end{lstlisting}

\subsubsection{SkillX}
The artifact is the complete hierarchical skill library supplied to the executor.

\paragraph{Checkpoint $K_1$.} The proposed library update contained no valid change, so no update was committed. The executor therefore retains the artifact from $K_0$.

\paragraph{Checkpoint $K_2$.} Skill version 1 contains 1,866 characters.
\begin{lstlisting}[style=artifactexample]
# SkillX hierarchical skill library

## Planning skills

### trim_overpriced_long
Activate when: observation.price > observation.public_fundamental; account.position > 0; risk_contract.max_order_size limits sell size
Relevant interfaces: observation.price, observation.public_fundamental, observation.public_signal, account.position, risk_contract
When the instrument trades materially above its public fundamental and the account is long, sell in size-capped increments to harvest overpricing and reduce exposure; otherwise hold when no edge exists.

## Functional skills

### overpricing_trim_procedure
Activate when: observation.price above observation.public_fundamental; account.position > 0; turnover and drawdown within risk limits
Relevant interfaces: observation.price, observation.public_fundamental, account.position, risk_contract.max_order_size, account.turnover_so_far, account.max_drawdown_so_far
Compute premium = (price - public_fundamental)/public_fundamental; if premium exceeds a small threshold and position > 0, place a limit sell for min(max_order_size, position) at or slightly below current price; otherwise submit no order.

## Atomic skills

### place_limit_sell_trim
Activate when: observation.price > observation.public_fundamental; account.position > 0
Relevant interfaces: observation.price, account.position, risk_contract.max_order_size
Submit a single limit sell order with quantity = min(max_order_size, current position) and limit price at or just below the current price to reduce an overpriced long position.

### compute_premium_pct
Activate when: observation.price and observation.public_fundamental available
Relevant interfaces: observation.price, observation.public_fundamental
Compute relative premium = (price - public_fundamental) / public_fundamental to decide whether the market is overpriced enough to warrant a trim.
\end{lstlisting}

\paragraph{Checkpoint $K_3$.} Skill version 2 contains 2,514 characters.
\begin{lstlisting}[style=artifactexample]
# SkillX hierarchical skill library

## Planning skills

### trim_overpriced_long
Activate when: observation.price > observation.public_fundamental; account.position > 0; observation.price - observation.public_signal > 0.10; account.turnover_so_far < risk_contract.max_turnover * 0.9
Relevant interfaces: observation.price, observation.public_fundamental, observation.public_signal, account.position, account.turnover_so_far, risk_contract.max_turnover
When the instrument trades above both its public fundamental and public signal (confirming genuine overpricing), and the account holds a long position, execute incremental limit sells capped by max_order_size to harvest mean-reversion profits while maintaining compliance with turnover and drawdown limits. Hold when no premium exists or when risk limits are approached.

## Functional skills

### overpricing_trim_procedure
Activate when: observation.price > observation.public_fundamental; account.position > 0; account.turnover_so_far + intended_quantity <= risk_contract.max_turnover; account.max_drawdown_so_far < risk_contract.max_drawdown
Relevant interfaces: observation.price, observation.public_fundamental, account.position, account.turnover_so_far, risk_contract.max_turnover, risk_contract.max_order_size
Execute the trim procedure: (1) compute premium_pct = (price - fundamental)/fundamental; (2) if premium_pct > 0.5%, calculate quantity = min(risk_contract.max_order_size, account.position); (3) verify turnover + quantity <= max_turnover; (4) submit limit sell at price - 0.01 to capture liquidity; (5) else submit no order.

## Atomic skills

### place_limit_sell_trim
Activate when: observation.price > observation.public_fundamental; account.position > 0; account.turnover_so_far < risk_contract.max_turnover
Relevant interfaces: observation.price, account.position, risk_contract.max_order_size, account.turnover_so_far, risk_contract.max_turnover
Submit a single limit sell order with quantity = min(risk_contract.max_order_size, account.position) and limit_price set 1 tick below the current price to ensure liquidity while reducing an overpriced long position.

### compute_premium_pct
Activate when: observation.price available; observation.public_fundamental available
Relevant interfaces: observation.price, observation.public_fundamental
Calculate the percentage premium as (observation.price - observation.public_fundamental) / observation.public_fundamental \xd7 100 to quantify the degree of overpricing before deciding on trim size.
\end{lstlisting}

\paragraph{Checkpoint $K_4$.} The proposed library update contained no valid change, so no update was committed. The executor therefore retains the artifact from $K_3$.

\paragraph{Checkpoint $K_5$.} The proposed library update contained no valid change, so no update was committed. The executor therefore retains the artifact from $K_4$.

\subsubsection{Trace2Skill}
The artifact is the complete skill document distilled from retained traces and supplied to the executor.

\paragraph{Checkpoint $K_1$.} Skill version 1 contains 2,271 characters.
\begin{lstlisting}[style=artifactexample]
# Skill: Trading in Valuation-Liquid Regime

## Regime

- `observable_feature_signature == "valuation:liquid"`

## Goal

Maximize risk-adjusted returns by being patient and selective. In this regime, the environment rewards low turnover and evidence-based trimming of overvaluation.

## Trading Rules

1. **Initial Step Behavior**
   - At the first decision step, do not trade immediately.
   - Hold the starting position and observe market development until valuation deviation clarifies.

2. **Overvaluation Trim Logic**
   - Continuously compare price vs `public_fundamental` and `public_signal`.
   - When price is meaningfully above both `public_fundamental` AND `public_signal` (approximately ~2.7% or more), execute a sell order to trim.
   - Sell quantity must respect risk limits: `qty <= max_order_size`.
   - Tag the order as `trim_overvalued`.
   - Only trim when overvaluation is clear and sustained across multiple steps, not a transient blip.

3. **Risk Management (must always hold)**
   - Never exceed `max_order_size` per order.
   - Respect `max_abs_position` at all times.
   - If `allow_short == false`, never let position go below 0.
   - Monitor drawdown closely; reduce trading activity if approaching `max_drawdown` limit.

4. **Hold Criteria**
   - If price is above fundamental but there is no clear/persistent overvaluation signal, hold.
   - Avoid acting on minor deviations (as a rule of thumb, < ~2% deviation should not trigger a trim).
   - Only act when the price-fundamental/signal gap is significant and persistent.

5. **Position Management**
   - Start from `initial_position` and manage position conservatively.
   - After trimming, reassess at subsequent observation points.
   - Do not buy when the market appears overvalued; wait for price correction toward fundamental/signal levels.

## Protected Behaviors (preserve)

- Strict adherence to `risk_contract` limits: `max_order_size`, `max_abs_position`, `max_drawdown`, and no shorting when `allow_short=false`.
- Patient initial observation at the first step before acting.
- Evidence-based trimming only on clear overvaluation, not minor deviations.
- Low-frequency, selective trading (avoid churn).
- Avoid counter-trend buying when overvalued relative to fundamental/signal.
\end{lstlisting}

\paragraph{Checkpoint $K_2$.} No valid patch was produced, so no update was committed. The executor therefore retains the artifact from $K_1$.

\paragraph{Checkpoint $K_3$.} The optimizer response could not be parsed, so no update was committed. The executor therefore retains the artifact from $K_2$.

\paragraph{Checkpoint $K_4$.} Skill version 2 contains 2,885 characters.
\begin{lstlisting}[style=artifactexample]
# Skill: Trading in Valuation-Liquid Regime

## Regime

- `observable_feature_signature == "valuation:liquid"`

## Goal

Maximize risk-adjusted returns by being patient and selective. In this regime, the environment rewards low turnover and evidence-based trimming of overvaluation.

## Trading Rules

1. **Initial Step Behavior**
   - At the first decision step, do not trade immediately.
   - Hold the starting position and observe market development until valuation deviation clarifies.

2. **Overvaluation Trim Logic**
   - Continuously compare price vs `public_fundamental` and `public_signal`.
   - When price is meaningfully above both `public_fundamental` AND `public_signal` (approximately ~2.7% or more), execute a sell order to trim. For persistent deviations between ~1.5% and ~2.7% sustained across many steps with a widening gap, a single conservative partial trim (one `max_order_size` clip, no repeat until deviation grows) is permitted. Sell quantity must respect `qty <= max_order_size`. Tag the order as `trim_overvalued`.
   - Sell quantity must respect risk limits: `qty <= max_order_size`.
   - Tag the order as `trim_overvalued`.
   - Only trim when overvaluation is clear and sustained across multiple steps, not a transient blip.
   - Execution: in liquid markets, price the sell limit at or slightly below the current price (e.g., a few ticks under mid) to ensure the trim actually fills; re-check fill status next step before assuming the position was reduced.

3. **Risk Management (must always hold)**
   - Never exceed `max_order_size` per order.
   - Respect `max_abs_position` at all times.
   - If `allow_short == false`, never let position go below 0.
   - Monitor drawdown closely; reduce trading activity if approaching `max_drawdown` limit.

4. **Hold Criteria**
   - If price is above fundamental but there is no clear/persistent overvaluation signal, hold.
   - Avoid acting on minor deviations (as a rule of thumb, < ~2% deviation should not trigger a trim).
   - Only act when the price-fundamental/signal gap is significant and persistent.

5. **Position Management**
   - Start from `initial_position` and manage position conservatively.
   - After trimming, reassess at subsequent observation points.
   - Do not buy when the market appears overvalued; wait for price correction toward fundamental/signal levels.

## Protected Behaviors (preserve)

- Strict adherence to `risk_contract` limits: `max_order_size`, `max_abs_position`, `max_drawdown`, and no shorting when `allow_short=false`.
- Patient initial observation at the first step before acting.
- Evidence-based trimming only on clear or codified persistent overvaluation, not transient blips.
- Low-frequency, selective trading; at most one conservative partial trim per escalation level, avoiding churn.
- Avoid counter-trend buying when overvalued relative to fundamental/signal.
\end{lstlisting}

\paragraph{Checkpoint $K_5$.} The optimizer response could not be parsed, so no update was committed. The executor therefore retains the artifact from $K_4$.

\subsubsection{SkillGrad}
The artifact is the complete layered skill package supplied to the executor.

\paragraph{Checkpoint $K_1$.} Skill version 1 contains 2,004 characters.
\begin{lstlisting}[style=artifactexample]
# SkillGrad layered skill package

## L1 routing metadata
Name: execution-verification
Description: Monitors post-trade execution to detect unfilled orders and triggers dynamic repricing when market trends bypass static limit orders, ensuring position alignment with directional thesis.

## L2 general guidance
## Execution Verification Protocol

**MANDATORY POST-TRADE CONFIRMATION LOOP**

After submitting any limit order, you must verify execution before proceeding with new strategy logic:

1. **Immediate Observation Check**: On the next observation cycle, explicitly verify that `position` size or `turnover` has changed in the direction of your order.
   - For buy orders: Confirm position increase or turnover > 0
   - For sell orders: Confirm position decrease or turnover > 0

2. **Staleness Detection**: If 2-3 consecutive observations show:
   - `turnover == 0` (no executions) OR
   - `position` unchanged despite active orders
   - While market price continues trending away from your limit price

   Then your orders are **stale** and the market has moved past your entry.

3. **Dynamic Repricing Rule**: When staleness is detected AND your directional thesis remains valid:
   - **Cancel** the stale limit orders immediately
   - **Reprice aggressively**: Use marketable limits (cross the spread) or adjust prices to within 0.5-1 tick of current bid/ask
   - For sell orders in an uptrend: Price at current ask minus small buffer (or use market orders if liquidity permits)
   - For buy orders in a downtrend: Price at current bid plus small buffer

**ANTI-PATTERN TO AVOID**: Do not submit limit orders and assume they will execute. In trending markets, static limits often result in zero position adjustment despite correct directional analysis.

**TREND RECOGNITION**: If price moves monotonically (e.g., three consecutive higher highs/lower lows) while your orders remain unfilled, immediately switch to aggressive repricing rather than waiting for mean reversion to your limit price.
\end{lstlisting}

\paragraph{Checkpoint $K_2$.} The proposed skill package was invalid or unchanged, so no update was committed. The executor therefore retains the artifact from $K_1$.

\paragraph{Checkpoint $K_3$.} The method committed an internal update, but the executor-visible skill document did not change. The executor therefore retains the artifact from $K_2$.

\paragraph{Checkpoint $K_4$.} The proposed momentum state was invalid, so no update was committed. The executor therefore retains the artifact from $K_3$.

\paragraph{Checkpoint $K_5$.} Skill version 3 contains 2,812 characters.
\begin{lstlisting}[style=artifactexample]
# SkillGrad layered skill package

## L1 routing metadata
Name: execution-verification
Description: Monitors post-trade execution to detect unfilled orders and triggers dynamic repricing when market trends bypass static limit orders, ensuring position alignment with directional thesis. Enforces uniform verification after every order submission and mandates marketable pricing in confirmed trends.

## L2 general guidance
## Execution Verification Protocol

**MANDATORY POST-TRADE CONFIRMATION LOOP**

After submitting any limit order, you must verify execution before proceeding with new strategy logic:

1. **Immediate Observation Check**: On the next observation cycle, explicitly verify that `position` size or `turnover` has changed in the direction of your order.
   - For buy orders: Confirm position increase or turnover > 0
   - For sell orders: Confirm position decrease or turnover > 0

2. **Staleness Detection**: If 2-3 consecutive observations show:
   - `turnover == 0` (no executions) OR
   - `position` unchanged despite active orders
   - While market price continues trending away from your limit price

   Then your orders are **stale** and the market has moved past your entry.

3. **Dynamic Repricing Rule**: When staleness is detected AND your directional thesis remains valid:
   - **Cancel** the stale limit orders immediately
   - **Reprice aggressively**: Use marketable limits (cross the spread) or adjust prices to within 0.5-1 tick of current bid/ask
   - For sell orders in an uptrend: Price at current ask minus small buffer (or use market orders if liquidity permits)
   - For buy orders in a downtrend: Price at current bid plus small buffer

**ANTI-PATTERN TO AVOID**: Do not submit limit orders and assume they will execute. In trending markets, static limits often result in zero position adjustment despite correct directional analysis.

**TREND RECOGNITION**: If price moves monotonically (e.g., three consecutive higher highs/lower lows) while your orders remain unfilled, immediately switch to aggressive repricing rather than waiting for mean reversion to your limit price.

**UNIFORM VERIFICATION RULE**: Apply this verification protocol after *every single order submission*, not just the first detected stale order. Incomplete position adjustment is a critical failure mode.

**CONFIRMED TREND EXECUTION RULE**: In a confirmed uptrend (defined as three or more consecutive higher prices), all sell orders intended for immediate execution MUST be priced at or below the current bid. Conversely, in a confirmed downtrend, all buy orders intended for immediate execution MUST be priced at or above the current ask. Never place passive sell limits above the market in an uptrend or passive buy limits below the market in a downtrend when the goal is immediate execution.
\end{lstlisting}

\section{Safety Implications of Persistent Self-Evolution}
\label{app:safety-discussion}
Persistent artifacts enable experience reuse but also allow a single erroneous update to propagate across many subsequent decisions. This creates a failure mode distinct from an isolated execution error: incorrect, overgeneralized, or obsolete instructions can persist across episodes and influence behavior beyond the context in which they were learned. Our retention analysis further shows that these failures tend to concentrate in a small number of evolution paths. Consequently, average performance can improve while individual paths suffer severe capability degradation.

The candidate-selection results reveal a second safety concern. Safeguards reduce harmful commits from $12.4\%$ to $6.2\%$, but increase missed improvements from $0.0\%$ to $16.5\%$. Meanwhile, agreement between selection-time validation and evaluation on unseen episodes remains close to $56\%$. Conservative acceptance thus reduces the frequency of observed regressions without improving the accuracy of candidate evaluation. This trade-off becomes especially critical after a rule change: rejecting an update may preserve an obsolete rule, whereas accepting it may impair capabilities that remain valid. The absence of reliable rule adaptation indicates that current methods cannot consistently resolve this distinction.

Safe self-evolution therefore requires explicit control over both update content and update persistence. Candidate instructions should retain their supporting evidence and the scope of affected capabilities. Updates should be evaluated on independent tasks that cover both preserved and revised behaviors, applied through bounded edits, and stored with versioning and rollback support. Deployment systems should also monitor tail losses and unsupported references to variables or rules absent from the agent's observation. This last recommendation is motivated by qualitative cases in which transferred skills improved the grounding of a decision explanation without improving its task score.

These results concern behavioral risks under controlled, non-adversarial task streams. EvoPathBench does not evaluate malicious artifact poisoning, prompt injection, privacy leakage, or deliberate reward manipulation. The experiments therefore demonstrate that persistent self-evolution introduces measurable update and retention risks, rather than establishing the safety of any evaluated method under open-world deployment.

\section{Future Work Directions}
\label{app:future-work}

Our results motivate several promising directions for advancing self-evolution beyond the methods and evaluation framework explored in this work.

\paragraph{Separating Preservation from Revision.}
Reliable evolution requires determining which knowledge should remain stable and which should be updated. Future methods could attach evidence, scope, and validity conditions to each learned rule. New feedback could then selectively revise only those rules whose assumptions are no longer supported, while preserving unrelated capabilities. Versioned artifacts and rollback mechanisms would provide additional safeguards when the consequences of an update become apparent only in later tasks.

\paragraph{Coordinating Artifacts and Parameter Updates.}
Persistent artifacts and parameter training offer distinct mechanisms for carrying experience forward. Future work could investigate when knowledge should be retained in an editable external artifact and when it should be distilled into model parameters. Joint methods might leverage artifacts for rapid, reversible adaptation, followed by selective parameter updates for capabilities supported by repeated evidence. Capability-level evaluation would help determine whether this coordination enhances transfer without exacerbating catastrophic forgetting.

\paragraph{Longer and Broader Evolution Paths.}
Future benchmarks could extend capability-level process evaluation to longer streams, multiple concurrent capability changes, and domains beyond calibrated markets. Such settings would enable the study of how agents manage several interacting rules, delayed feedback, and changes whose boundaries are not explicitly marked. They would also clarify which principles of artifact evolution transfer across environments and which depend on the structure of a particular task domain.
\end{document}